\documentclass[12pt,halfline,a4paper]{ouparticle}

\usepackage{enumerate}
\usepackage{amssymb}
\usepackage{amsmath}
\usepackage{graphicx,subfigure}
\usepackage{epstopdf}
\usepackage{multirow}
\usepackage{algorithm}
\usepackage[noend]{algpseudocode}
\usepackage[numbers]{natbib}
\usepackage{algorithmicx}
\usepackage{dirtytalk}
\usepackage{scrextend}
\usepackage{xcolor,colortbl}
\usepackage{pbox}
\usepackage{pdfpages}
\usepackage{enumitem}
\usepackage{pdflscape}
\usepackage{marginnote}
\usepackage{caption}
\usepackage[T1]{fontenc}
\usepackage{lipsum}
\usepackage[T1]{fontenc}
\usepackage[utf8]{inputenc}
\usepackage{authblk}
\usepackage{changepage}
\usepackage{hyperref}
\hypersetup{
        colorlinks=true,
        linkcolor=blue,
        citecolor = blue,
        anchorcolor = blue,
        urlcolor=blue,
        linktoc=page,
}

\title{Deep Face Image Retrieval: a Comparative Study with Dictionary Learning}
\author[1]{Ahmad S. Tarawneh}
\author[2]{Ahmad B. A. Hassanat}
\author[3]{Ceyhun Celik}
\author[1]{Dmitry Chetverikov}
\author[4]{M. Sohel Rahman}
\author[1]{Chaman  Verma}

\affil[1]{Department of Algorithms and Their Applications , \textit{E\"{o}tv\"{o}s Lor\'{a}nd University}, Budapest, Hungary}
\affil[2]{Department of Information Technology, Mutah University, Karak, Jordan}
\affil[3]{Department of Computer Engineering, Gazi University, Ankara, Turkey}
\affil[4]{Department of CSE, BUET, ECE Building, West Palasi, Dhaka 1205, Bangladesh}

\begin{document}

\abstract{Facial image retrieval is a challenging task since faces have many similar features (areas), which makes it difficult for the retrieval systems to distinguish faces of different people. With the advent of deep learning, deep networks are often applied to extract powerful features that are used in many areas of computer vision. This paper investigates the application of different deep learning models for face image retrieval, namely, Alexlayer6, Alexlayer7, VGG16layer6, VGG16layer7, VGG19layer6, and VGG19layer7, with two types of dictionary learning techniques, namely $K$-means and $K$-SVD. We also investigate some coefficient learning techniques such as the Homotopy, Lasso, Elastic Net and SSF and their effect on the face retrieval system. The comparative results of the experiments conducted on three standard face image datasets show that the best performers for face image retrieval are Alexlayer7 with $K$-means and SSF, Alexlayer6 with $K$-SVD and SSF, and Alexlayer6 with $K$-means and SSF. The APR and ARR of these methods were further compared to some of the state of the art methods based on local descriptors. The experimental results show that deep learning outperforms most of those methods and therefore can be recommended for use in practice of face image retrieval}


\maketitle

\section{Introduction}
Both facial recognition and facial image retrieval (FIR) are important problems of computer vision and image processing. The main difference between the two problems is that in the former we try to identify or verify a person from a digital image of person's face, while in the latter we need to retrieve $N$ facial images that are relevant to a query face image \cite{gudivada1995content}\cite{hassanat2016fusion}\cite{tarawneh2018stability}\cite{tarawneh2018detailed}. Similar to face recognition systems, a facial image retrieval (FIR) system works by extracting useful features to be used in the retrieval process. The focus of this study is the latter which is the more difficult problem of the two. The main difference between both systems is that in retrieval systems you need to retrieve N facial images which are relevant to a query face image.

Facial image retrieval has been under investigation for years and a lot of methods have been employed to develop this field. FIR can be seen as a content-based image retrieval (CBIR) problem that focuses on extracting and retrieving facial images rather than any other content. The major challenge in FIR is that the features of human faces can change due to various expressions, poses, hair style as well as through artificial manipulations (e.g., tattoo or painting on the face). All these factors make it difficult for a system to extract stable and robust features to be used in face recognition and FIR systems. 

A number of algorithms have been developed to enhance the accuracy of FIR. For example, Chen et al. \cite{chen2013scalable} proposed enhancements for FIR by extracting semantic information and used the datasets LFW \cite{huang2008labeled} and Pubfig \cite{kumar2009attribute} to test the methods. They used small number of query images (120) and their results on LFW were worse than that on Pubfig. Notably, the random selection of query images in their experiments makes it difficult to make a fair comparison with their work. 
Local Gradient Hexa Pattern (LGHP) \cite{chakraborty2018local} was also proposed to extract descriptive features for face recognition and retrieval. The proposed features were invariant to pose and  illumination variation. A number of datasets were used to test the proposed method including two challenging datasets, namely, Yale-B \cite{georghiades2001few,lee2005acquiring} and LFW. However, the results on LFW were not satisfactory as the highest retrieval rate achieved was around 13.5\% on the first retrieved image. 

In recent times, Deep Learning techniques are widely being used to solve many problems of computer vision (e.g.,  \cite{tarawneh2018pilot}\cite{saritha2018content}\cite{tzelepi2018deep}
\cite{khatami2018sequential}\cite{pang2018novel}). Although Deep learning is preferred to Sparse Representations (SR) to improve retrieval accuracy in content-based image retrieval (CBIR) problems \cite{cheng2016learning}\cite{lei2016learning}\cite{kappeler2016video}\cite{gao2015single}, these approaches can also be combined for the same purpose \cite{zhao2015heterogeneous}\cite{liu2016robust}\cite{dong2016image}\cite{goh2014learning}. This was the main motivation of the current research. In particular, our current study conducts extensive experiments to find out the best combination of these two approaches with a goal to improve the performance of FIR systems and thereby advance the state of the art.  

The main contribution of this paper is as follows. We make a fusion of DL and SR to make the best out of the two approaches. 

In addition, we combine deep features (DF) with a coefficient learning method, namely, separable surrogate function (SSF), which, to the best of our knowledge, has not been used before in the literature in the context of FIR. We have extracted different DFs from the most popular deep network models, namely, Alexnet \cite{krizhevsky2012imagenet} and VGG \cite{simonyan2014very} to ensure a robust FIR system. To evaluate our approach, we have conducted a thorough experimental analysis using two of the most challenging datasets, namely, LFW, YAL-B and Gallagher. Our experiments show that combining DF with SR enhances FIR considerably.

\section{Methods}

Several approaches exist to use the deep learning. For example, Convolutional Neural Networks (CNN) can be employed with full-training on a new dataset from scratch. However, this approach requires a large amount of data since a lot of parameters of CNN need to be tuned. Another approach is based on fine-tuning a specific layer or several of them and changing/updating their parameters to fit the new data. The third, and perhaps the most common, approach uses a pre-trained CNN model that has already been adequately trained using a huge dataset that contains millions of images (e.g., Imagenet database \cite{deng2009imagenet}), albeit with a different goal: in this case, the pre-trained CNN is used to extract descriptive features to be exploited for different tasks such as face image retrieval and recognition. In this work we opt for the third approach and make an effort to extract deep features from different layers of the pre-trained models, such as AlexNet \cite{krizhevsky2012imagenet}, VGG-16 and VGG-19 \cite{simonyan2014very}. 

Each of the models we use provides a 4096 dimensional feature vector to represent the content of each image. We also have used different fully-connected layers (FCL), FCL-6 and FCL-7, to extract the features from each model. In other words, we investigate the use of Alexnet with FCL-6 (Alexlayer6), Alexnet with FCL-7 (Alexlayer7), VGG-16 with FCL-6 (VGG16layer6), VGG-16 with FCL-7 (VGG16layer7), VGG-19 with FCL-6 (VGG19layer6), and VGG-19 with FCL-7 (VGG19layer7) to experimentally analyze which one is preferable for face image retrieval. Furthermore, we use two types of dictionary learning methods, namely, $K$-means and $K$-singular value decomposition ($K$-SVD) \cite{aharon2006k}, while Homotopy, Lasso, Elastic Net and SSF \cite{de1993relation}\cite{celik2017content} are used as coefficient learning techniques with each method.

\begin{figure}[th]
		\centering
		\includegraphics[width=1\linewidth]{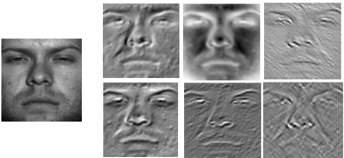}
		\caption[]{Left: an image from YAL-B dataset. Rest: six representations of the image provided Alexnet layers.}	
		\label{fig:facepresntation}
	\end{figure}

Both Alexnet and VGG networks extract the features in almost the same way. The input image is provided to the input layer of each model, and then it is processed through the different convolutional layers to obtain different representations using several filters. Figure \ref{fig:facepresntation} shows how a single face image is represented by different CNN layers.

\subsection{Sparse Representation}	\label{sec:SR}
	
Based on the principle of sparsity, any vector could be represented with a few non-zero element according to a base. If this idea is applied to the problem of extracting meaningful information from a bunch of vectors, all of the vectors will be represented with a simple coefficients on the same base. Solving the problem is thus simplified with Sparse Representations (SRs) of vectors. Although this technique has been applied in signal processing for many years, during the last two decades it has also been used for solving computer vision problems, such as, image retrieval, image denoising, and image classification \cite{Wright2009}. The goal of SR is achieved by solving following problem:
	
	\begin{equation}
	\min_{\alpha\in R_{}^n} \frac{1}{2} \|x-D\alpha\|_{2}^{2}+\lambda \|\alpha\|_{p}.
	\label{SR1}
	\end{equation}
\noindent	
Here $x$ is the signal, $D$ the dictionary and $\alpha$ the sparse coefficient of signal $x$; $p$ could be any value from $[0,\infty]$. Solving this problem is realized in two steps, namely, Dictionary Learning (DL) and Coefficient Learning (CL). The base is built in the DL step and the sparse coefficients of vectors are obtained in the CL step. In the literature, DL algorithms are categorized as offline or online \cite{Rubinstein2010,Coates2011}. Offline DL algorithms, such as, $K$-Means algorithm, build the dictionary without any help of sparse coefficients. On the other hand, online algorithms, such as, $K$-Singular Value Decomposition ($K$-SVD), incorporate the sparse coefficients in the dictionary building process. On line DL algorithms thus make use of CL algorithms to build the dictionary as follows. First, they obtain the sparse coefficients with a random dictionary. Then, the dictionary is trained with the obtained sparse coefficients. These two steps are repeated iteratively, until a stopping criteria is achieved. 

Unlike the DL step, there are many solutions for the CL step \cite{Bach2012}. Solutions like Homotopy, Lasso, Elastic Net are generally greedy approaches \cite{Zibulevsky2010}. Unfortunately however these greedy solutions are inefficient for high-dimensional problems \cite{Zibulevsky2010}. On the other hand, iterative-shrinkage algorithms, such as, SSF and Parallel Coordinate Descent (PCD) are reported to produce effective solutions to high-dimensional problems like image retrieval \cite{celik2017content}. Here, instead of solving the SR problem (Eq. \ref{SR1}), a surrogate function is  applied to obtain sparse coefficients as follows:
 
 	\begin{equation}
 	f_{}^*(\alpha)=\frac{1}{2}\|x-D\alpha\|_2^2+\lambda1_{}^Tp(\alpha)	+\frac{c}{2}\|\alpha-\alpha_0\|_2^2-\frac{1}{2}\|D\alpha-D\alpha_0\|_2^2	
 	\label{SR2}
 	\end{equation}
 	
 	This surrogate function is obtained with the following additional term:
 	
 	\begin{equation}
 	d(\alpha,\alpha_0)=\frac{c}{2}\|\alpha-\alpha_0\|_2^2-\frac{1}{2}\|D\alpha-D\alpha_0\|_2^2
 	\label{SR3}
 	\end{equation} 
 	
The setting of the parameter $c$ should guarantee that the function $d$ is strictly convex. The surrogate function takes place of the minimization term. Thus, the task of obtaining sparse coefficient becomes much simpler and can be solved efficiently, since the minimization term $\|D\alpha_0\|_2^2$ is nonlinear\cite{Zibulevsky2010}.
	
In this study, two traditional DL algorithms, $K$-Means as an offline approach and $K$-SVD as an online approach, are used to build the dictionary. Then, sparse coefficients are obtained by greedy approaches (Homotopy, Lasso, Elastic Net) as well as by an iterative shrinkage algorithm (SSF).

\subsection{Datasets}
In our experiments, we have used three of the most common and challenging face image benchmark datasets, namely, the Cropped Extended Yale B (Yale-B) \cite{georghiades2001few} \cite{lee2005acquiring}, the Labeled Faces in the Wild (LFW) \cite{huang2008labeled} and Gallagher \cite{gallagher2008clothing}. The YAL-B dataset consists of 38 classes (different subjects) each having 65 images. The LFW dataset consists of 5749 subjects each having different number of images ranging from 1 to 530. Since most of the subjects in the LFW dataset have only one image, following \cite{chakraborty2018local} we have used a subset of LFW by choosing only the subjects that contain at least 20 images. Thus, we were left with only 62 subjects each having different number (at least 20) of images.

\section{Results}

All the aforementioned methods for deep feature extraction and different dictionary learning approaches have been implemented using Matlab 2018b and run in a machine with NVIDIA GeForce GTX 1080 GPU having windows 10 as the OS. We have run the methods on all three datasets to extract the features from the face images. In our experiments, 10 images of each subject from the datasets have been used as query images, while we have used the rest for training. Tables 1, 2 and 3 show the Mean Average Precision (MAP) of the face image retrieval from the Yale-B dataset. 

\begin{table}[ht]
\centering
\caption{MAP results of all the face images retrieved from the Yale-B dataset. The best results have been highlighted using boldface fonts.}
\label{Table1}
\begin{tabular}{ccccc|cccc}
\textbf{} & \multicolumn{4}{c}{\textbf{38-D $K$-Means}} & \multicolumn{4}{c}{\textbf{38-D $K$-SVD}} \\
\textbf{} & \textbf{Homo} & \textbf{Lasso} & \textbf{\begin{tabular}[c]{@{}c@{}}Elastic\\  Net\end{tabular}} & \textbf{SSF} & \textbf{Homo} & \textbf{Lasso} & \textbf{\begin{tabular}[c]{@{}c@{}}Elastic\\  Net\end{tabular}} & \textbf{SSF} \\ \hline
\textbf{Alexlayer6} & 0.39 & 0.24 & 0.08 & \textbf{0.48} & 0.3 & 0.26 & 0.1 & 0.45 \\
\textbf{Alexlayer7} & 0.37 & 0.26 & 0.08 & \textbf{0.49} & 0.37 & 0.31 & 0.14 & \textbf{0.49} \\
\textbf{VGG16layer6} & 0.34 & 0.19 & 0.03 & \textbf{0.42} & 0.25 & 0.21 & 0.04 & 0.39 \\
\textbf{VGG16layer7} & 0.32 & 0.19 & 0.06 & \textbf{0.41} & 0.24 & 0.18 & 0.08 & 0.38 \\
\textbf{VGG19layer6} & 0.29 & 0.17 & 0.05 & \textbf{0.39} & 0.19 & 0.15 & 0.05 & 0.34 \\
\textbf{VGG19layer7} & 0.3 & 0.23 & 0.05 & \textbf{0.39} & 0.22 & 0.18 & 0.07 & 0.35
\end{tabular}
\end{table}

\begin{table}[ht]
\caption{MAP results of the first 10 face images retrieved from the Yale-B dataset. The best results have been highlighted using boldface fonts.}
\label{Table2}
\begin{tabular}{ccccc|cccc}
\textbf{} & \multicolumn{4}{c}{\textbf{38-D $K$-Means}} & \multicolumn{4}{c}{\textbf{38-D $K$-SVD}} \\
\textbf{} & \textbf{Homo} & \textbf{Lasso} & \textbf{\begin{tabular}[c]{@{}c@{}}Elastic\\  Net\end{tabular}} & \textbf{SSF} & \textbf{Homo} & \textbf{Lasso} & \textbf{\begin{tabular}[c]{@{}c@{}}Elastic\\  Net\end{tabular}} & \textbf{SSF} \\ \hline
\textbf{Alexlayer6} & 0.66 & 0.35 & 0.12 & \textbf{0.79} & 0.5 & 0.41 & 0.17 & 0.78 \\
\textbf{Alexlayer7} & 0.67 & 0.4 & 0.15 & \textbf{0.8} & 0.55 & 0.44 & 0.22 & \textbf{0.8} \\
\textbf{VGG16layer6} & 0.62 & 0.29 & 0.03 & \textbf{0.76} & 0.42 & 0.31 & 0.04 & 0.73 \\
\textbf{VGG16layer7} & 0.61 & 0.29 & 0.08 & \textbf{0.75} & 0.39 & 0.26 & 0.12 & 0.71 \\
\textbf{VGG19layer6} & 0.55 & 0.25 & 0.06 & \textbf{0.73} & 0.31 & 0.22 & 0.07 & 0.67 \\
\textbf{VGG19layer7} & 0.57 & 0.34 & 0.07 & \textbf{0.72} & 0.38 & 0.29 & 0.09 & 0.66
\end{tabular}
\end{table}

\begin{table}[ht]
\caption{MAP results of the first 5 face images retrieved from the Yale-B dataset. The best results have been highlighted using boldface fonts.}
\label{Table3}
\begin{tabular}{ccccc|cccc}
\textbf{} & \multicolumn{4}{c}{\textbf{38-D $K$-Means}} & \multicolumn{4}{c}{\textbf{38-D $K$-SVD}} \\
\textbf{} & \textbf{Homo} & \textbf{Lasso} & \textbf{\begin{tabular}[c]{@{}c@{}}Elastic\\  Net\end{tabular}} & \textbf{SSF} & \textbf{Homo} & \textbf{Lasso} & \textbf{\begin{tabular}[c]{@{}c@{}}Elastic\\  Net\end{tabular}} & \textbf{SSF} \\ \hline
\textbf{Alexlayer6} & 0.74 & 0.39 & 0.13 & \textbf{0.86} & 0.56 & 0.44 & 0.18 & 0.85 \\
\textbf{Alexlayer7} & 0.74 & 0.45 & 0.16 & \textbf{0.86} & 0.61 & 0.48 & 0.24 & \textbf{0.86} \\
\textbf{VGG16layer6} & 0.7 & 0.32 & 0.03 & \textbf{0.83} & 0.48 & 0.34 & 0.05 & 0.81 \\
\textbf{VGG16layer7} & 0.69 & 0.32 & 0.08 & \textbf{0.83} & 0.45 & 0.29 & 0.13 & 0.8 \\
\textbf{VGG19layer6} & 0.64 & 0.29 & 0.06 & \textbf{0.82} & 0.37 & 0.25 & 0.07 & 0.76 \\
\textbf{VGG19layer7} & 0.66 & 0.37 & 0.08 & \textbf{0.8} & 0.45 & 0.32 & 0.1 & 0.75
\end{tabular}
\end{table}
As is evident from the results on the Yale-B dataset, SSF is quite convincingly the best performer among the coefficient learning techniques and among the two dictionary learning techniques, $K$-Means performs slightly better. It is also evident that AlexNet features have a slight edge over the the VGG-16 and VGG-19 features.
\newpage
Now we focus on the results of the LFW dataset which have been presented in Tables 4, 5 and 6. 

\begin{table}[ht]
\caption{MAP results of all the face images retrieved from LFW dataset. The best results have been highlighted using boldface fonts.}
\label{Table4}
\begin{tabular}{ccccc|cccc}
\textbf{} & \multicolumn{4}{c}{\textbf{38-D $K$-Means}} & \multicolumn{4}{c}{\textbf{38-D $K$-SVD}} \\
\textbf{} & \textbf{Homo} & \textbf{Lasso} & \textbf{\begin{tabular}[c]{@{}c@{}}Elastic\\  Net\end{tabular}} & \textbf{SSF} & \textbf{Homo} & \textbf{Lasso} & \textbf{\begin{tabular}[c]{@{}c@{}}Elastic\\  Net\end{tabular}} & \textbf{SSF} \\ \hline
\textbf{Alexlayer6} & 0.11 & 0.12 & 0.07 & 0.16 & 0.13 & 0.1 & 0.08 & \textbf{0.17} \\
\textbf{Alexlayer7} & 0.08 & 0.08 & 0.06 & \textbf{0.13} & 0.12 & 0.1 & 0.09 & \textbf{0.13} \\
\textbf{VGG16layer6} & 0.06 & 0.06 & 0.05 & \textbf{0.1} & 0.07 & 0.06 & 0.06 & \textbf{0.1} \\
\textbf{VGG16layer7} & 0.05 & 0.07 & 0.07 & \textbf{0.08} & 0.08 & 0.06 & 0.06 & 0.08 \\
\textbf{VGG19layer6} & 0.06 & 0.06 & 0.05 & \textbf{0.1} & 0.07 & 0.06 & 0.06 & 0.09 \\
\textbf{VGG19layer7} & 0.05 & 0.06 & 0.07 & \textbf{0.08} & 0.07 & 0.06 & 0.06 & 0.08
\end{tabular}
\end{table}

\begin{table}[]
\caption{MAP results of the first 10 face images retrieved from LFW dataset. The best results have been highlighted using boldface fonts.}
\label{Table5}
\begin{tabular}{ccccc|cccc}
\textbf{} & \multicolumn{4}{c}{\textbf{38-D $K$-Means}} & \multicolumn{4}{c}{\textbf{38-D $K$-SVD}} \\
\textbf{} & \textbf{Homo} & \textbf{Lasso} & \textbf{\begin{tabular}[c]{@{}c@{}}Elastic\\  Net\end{tabular}} & \textbf{SSF} & \textbf{Homo} & \textbf{Lasso} & \textbf{\begin{tabular}[c]{@{}c@{}}Elastic\\  Net\end{tabular}} & \textbf{SSF} \\ \hline
\textbf{Alexlayer6} & 0.15 & 0.13 & 0.06 & 0.2 & 0.15 & 0.11 & 0.07 & \textbf{0.21} \\
\textbf{Alexlayer7} & 0.1 & 0.09 & 0.05 & 0.16 & 0.14 & 0.11 & 0.08 & \textbf{0.17} \\
\textbf{VGG16layer6} & 0.08 & 0.06 & 0.04 & \textbf{0.12} & 0.08 & 0.06 & 0.05 & \textbf{0.12} \\
\textbf{VGG16layer7} & 0.06 & 0.08 & 0.07 & \textbf{0.1} & 0.09 & 0.07 & 0.07 & \textbf{0.1} \\
\textbf{VGG19layer6} & 0.08 & 0.07 & 0.05 & \textbf{0.12} & 0.08 & 0.07 & 0.05 & 0.11 \\
\textbf{VGG19layer7} & 0.06 & 0.07 & 0.06 & \textbf{0.1} & 0.09 & 0.07 & 0.06 & \textbf{0.1}
\end{tabular}
\end{table}

\begin{table}[ht]
\caption{MAP results of the first 5 face images retrieved from LFW dataset. The best results have been highlighted using boldface fonts.}
\label{Table6}
\begin{tabular}{ccccc|cccc}
\textbf{} & \multicolumn{4}{c}{\textbf{38-D $K$-Means}} & \multicolumn{4}{c}{\textbf{38-D $K$-SVD}} \\
\textbf{} & \textbf{Homo} & \textbf{Lasso} & \textbf{\begin{tabular}[c]{@{}c@{}}Elastic\\  Net\end{tabular}} & \textbf{SSF} & \textbf{Homo} & \textbf{Lasso} & \textbf{\begin{tabular}[c]{@{}c@{}}Elastic\\  Net\end{tabular}} & \textbf{SSF} \\ \hline
\textbf{Alexlayer6} & 0.17 & 0.14 & 0.07 & 0.23 & 0.17 & 0.12 & 0.08 & \textbf{0.24} \\
\textbf{Alexlayer7} & 0.12 & 0.1 & 0.05 & 0.19 & 0.15 & 0.12 & 0.08 & \textbf{0.19} \\
\textbf{VGG16layer6} & 0.08 & 0.07 & 0.04 & 0.13 & 0.08 & 0.07 & 0.06 & \textbf{0.14} \\
\textbf{VGG16layer7} & 0.07 & 0.09 & 0.07 & 0.11 & 0.1 & 0.07 & 0.07 & \textbf{0.12} \\
\textbf{VGG19layer6} & 0.09 & 0.07 & 0.05 & \textbf{0.13} & 0.09 & 0.07 & 0.05 & \textbf{0.13} \\
\textbf{VGG19layer7} & 0.07 & 0.07 & 0.07 & \textbf{0.11} & 0.1 & 0.07 & 0.06 & \textbf{0.11}
\end{tabular}
\end{table}
On LFW dataset, we get mixed results with respect to the dictionary learning approaches. As we can see, considering all retrievals, $K$-Means still has a small edge over $K$-SVD but that quickly diminishes as we become selective: for the first 5 images retrieved, the latter in fact shows better performance than the former. SSF however, consistently achieves the best performance as it did on Yale-B dataset. AlexNet features are still found to be superior here.

\begin{table}[ht]
\caption{MAP results of all the face images retrieved from Gallagher dataset. The best results have been highlighted using boldface fonts.}
\label{Table7}
\begin{tabular}{ccccc|cccc}
\textbf{} & \multicolumn{4}{c}{\textbf{38-D $K$-Means}} & \multicolumn{4}{c}{\textbf{38-D $K$-SVD}} \\
\textbf{} & \textbf{Homo} & \textbf{Lasso} & \textbf{\begin{tabular}[c]{@{}c@{}}Elastic\\  Net\end{tabular}} & \textbf{SSF} & \textbf{Homo} & \textbf{Lasso} & \textbf{\begin{tabular}[c]{@{}c@{}}Elastic\\  Net\end{tabular}} & \textbf{SSF} \\ \hline
\textbf{Alexlayer6} & 0.4 & 0.42 & 0.24 & \textbf{0.44} & 0.39 & 0.38 & 0.34 & 0.43 \\
\textbf{Alexlayer7} & 0.34 & 0.36 & 0.25 & 0.37 & 0.37 & 0.36 & 0.34 & \textbf{0.39} \\
\textbf{VGG16layer6} & 0.38 & 0.41 & 0.35 & 0.42 & \textbf{0.43} & \textbf{0.43} & 0.39 & \textbf{0.42} \\
\textbf{VGG16layer7} & 0.34 & 0.37 & 0.36 & 0.37 & 0.37 & 0.37 & 0.38 & \textbf{0.38} \\
\textbf{VGG19layer6} & 0.36 & 0.38 & 0.33 & 0.41 & 0.41 & 0.41 & 0.38 & \textbf{0.43} \\
\textbf{VGG19layer7} & 0.29 & 0.33 & 0.33 & 0.34 & 0.36 & 0.35 & 0.35 & \textbf{0.38}
\end{tabular}
\end{table}

\begin{table}[ht]
\caption{MAP results of the first 10 face images retrieved from Gallagher dataset. The best results have been highlighted using boldface fonts.}
\label{Table8}
\begin{tabular}{ccccc|cccc}
\textbf{} & \multicolumn{4}{c}{\textbf{38-D $K$-Means}} & \multicolumn{4}{c}{\textbf{38-D $K$-SVD}} \\
\textbf{} & \textbf{Homo} & \textbf{Lasso} & \textbf{\begin{tabular}[c]{@{}c@{}}Elastic\\  Net\end{tabular}} & \textbf{SSF} & \textbf{Homo} & \textbf{Lasso} & \textbf{\begin{tabular}[c]{@{}c@{}}Elastic\\  Net\end{tabular}} & \textbf{SSF} \\ \hline
\textbf{Alexlayer6} & 0.52 & 0.49 & 0.19 & \textbf{0.56} & 0.47 & 0.44 & 0.31 & 0.51 \\
\textbf{Alexlayer7} & 0.41 & 0.45 & 0.21 & 0.45 & 0.44 & 0.42 & 0.34 & \textbf{0.49} \\
\textbf{VGG16layer6} & 0.46 & 0.47 & 0.29 & \textbf{0.5} & 0.47 & 0.46 & 0.38 & 0.48 \\
\textbf{VGG16layer7} & 0.4 & 0.43 & 0.41 & 0.43 & 0.42 & 0.42 & 0.39 & \textbf{0.45} \\
\textbf{VGG19layer6} & 0.44 & 0.42 & 0.23 & \textbf{0.49} & 0.43 & 0.42 & 0.43 & 0.47 \\
\textbf{VGG19layer7} & 0.35 & 0.38 & 0.34 & 0.4 & 0.4 & 0.39 & 0.37 & \textbf{0.44}
\end{tabular}
\end{table}

\begin{table}[ht]
\caption{MAP results of the first 5 face images retrieved from Gallagher dataset. The best results have been highlighted using boldface fonts.}
\label{Table9}
\begin{tabular}{ccccc|cccc}
\textbf{} & \multicolumn{4}{c}{\textbf{38-D $K$-Means}} & \multicolumn{4}{c}{\textbf{38-D $K$-SVD}} \\
\textbf{} & \textbf{Homo} & \textbf{Lasso} & \textbf{\begin{tabular}[c]{@{}c@{}}Elastic\\  Net\end{tabular}} & \textbf{SSF} & \textbf{Homo} & \textbf{Lasso} & \textbf{\begin{tabular}[c]{@{}c@{}}Elastic\\  Net\end{tabular}} & \textbf{SSF} \\ \hline
\textbf{Alexlayer6} & 0.55 & 0.5 & 0.2 & \textbf{0.58} & 0.49 & 0.45 & 0.3 & 0.52 \\
\textbf{Alexlayer7} & 0.42 & 0.47 & 0.23 & 0.46 & 0.46 & 0.43 & 0.34 & \textbf{0.52} \\
\textbf{VGG16layer7} & 0.41 & 0.44 & 0.4 & 0.45 & 0.43 & 0.43 & 0.4 & \textbf{0.46} \\
\textbf{VGG19layer6} & 0.45 & 0.43 & 0.24 & \textbf{0.5} & 0.44 & 0.42 & 0.43 & 0.47 \\
\textbf{VGG19layer7} & 0.36 & 0.39 & 0.33 & 0.41 & 0.42 & 0.4 & 0.38 & \textbf{0.45}
\end{tabular}
\end{table}

The best results on Gallagher dataset switch between $K$-Means and $K$-SVD. However, SSF is still the best with AlexNet features. The best MAP on the whole dataset is 44\% using AlexNet layer 6 with SSF and $K$-Means.

Noticeably, AlexNet performs better than VGG across all three datasets. Although it is unlikely for AlexNet to achieve higher precision than VGG, in practical applications the information density provided by AlexNet is better than that of VGG, and therefore, AlexNet provides better utilization for its parameter space than VGG, particularly, for face image retrieval systems. This claim is also supported by \cite{canziani2016analysis}
\clearpage
According to the above experiments, we can say with certain confidence that AlexNet layers 6 and 7 have performed the best in terms of MAP, particularly, when using $K$-SVD dictionary learning with SSF coefficient learning. Therefore, we compare their results with the state-of-the-art face image retrieval methods. Since local descriptors explore higher order derivative space and tend to achieve better results under pose, expression, light, and
illumination variations \cite{chakraborty2018local}, we consider four of popular descriptors, namely, local gradient hexa pattern (LGHP) \cite{chakraborty2018local}, local derivative pattern (LDP) \cite{zhang2010local}, local tetra pattern (LTrP) \cite{murala2012local} and local vector pattern (LVP) \cite{fan2014novel}. These methods calculated the Average Precision of Retrieval (APR) of each dataset using the first retrieved face image, the first 5 retrieved face images, and the first (8 or 10) retrieved face images; therefore we also calculated the APR of AlexNet FC6 and AlexNet FC7 similarly. Tables 10 reports the results of this comparative analysis. As can be seen from Tables 10, the combination of AlexNet features with SSF enhances the results as it outperforms all four methods under consideration on all of the datasets. This significant increase of the APR can be attributed to the power of the deep features as compared to hand-crafted features, in general. 

However, in certain cases/datasets, Deep features might not be the magic tool for computer vision tasks. This is evident from another analysis reported in Table 11, where the Average Recall of Retrieval (ARR) have been reported instead of APR. Here we can see that while the ARR results in Yale-B and LFW datasets are dominated by our combined method involving deep features, on the Gallagher dataset, ARRs obtained by the LGHP, LDP, and LVP are found to be better.

\begin{table}[ht]
\centering
\caption{APR results of the selected deep learning methods compared to that of some local descriptors. APR@n implies APR result on the first $n$ retrieved face images.}
\label{Table10}
\begin{tabular}{cccc}
\hline
\multirow{2}{*}{\textbf{Method}} & \multicolumn{3}{c}{\textbf{YaleB (\%)}} \\ \cline{2-4} 
 & \textbf{APR@1} & \textbf{APR@5} & \textbf{APR@8} \\ \hline
Alexnet layer 7+ SSF + Kmeans & \textbf{90.5} & \textbf{81.8} & \textbf{73.6} \\
LGHP & 85 & 65 & 55 \\
LDP & 55 & 42 & 40 \\
LTrP & 37 & 30 & 25 \\
LVP & 81 & 60 & 50 \\ \hline
 & \multicolumn{3}{c}{\textbf{LFW (\%)}} \\ \cline{2-4} 
 & \textbf{APR@1} & \textbf{APR@5} & \textbf{APR@10} \\ \hline
Alexnet layer 6+ SSF + KSVD & \textbf{29.68} & \textbf{20.32} & \textbf{16.81} \\
LGHP & 13.5 & 7.5 & 6 \\
LDP & 8.5 & 5.5 & 4.5 \\
LTrP & 6.5 & 4.3 & 4 \\
LVP & 11 & 6.5 & 5.5 \\ \hline
\textbf{} & \multicolumn{3}{c}{\textbf{Gallagher(\%)}} \\ \cline{2-4} 
\textbf{} & \textbf{APR@1} & \textbf{APR@3} & \textbf{APR@5} \\ \hline
Alexnet layer 6+ SSF + Kmeans & \textbf{59.1} & \textbf{57.8} & \textbf{56.3} \\
LGHP & 51.8 & 39 & 30 \\
LDP & 36.5 & 27 & 25 \\
LTrP & 23.1 & 17 & 15 \\
LVP & 41.5 & 33 & 27 \\ \hline
\end{tabular}
\end{table}

\begin{table}[ht]
\centering

\caption{ARR results of the selected deep learning methods compared to that of some local descriptors. ARR@n implies ARR result on the first $n$ retrieved face images.}
\label{Table11}
\begin{tabular}{cccc}
\hline
\multirow{2}{*}{\textbf{Method}} & \multicolumn{3}{c}{\textbf{YaleB (\%)}} \\ \cline{2-4} 
 & \textbf{ARR@1} & \textbf{ARR@5} & \textbf{ARR@8} \\ \hline
Alexnet layer 7+ SSF + Kmeans & \textbf{1.7} & \textbf{7.4} & \textbf{12.6} \\
LGHP & 1.4 & 5 & 6.2 \\
LDP & 1 & 3 & 4.5 \\
LTrP & 0.5 & 2 & 2.9 \\
LVP & 1 & 4.9 & 6 \\ \hline
 & \multicolumn{3}{c}{\textbf{LFW (\%)}} \\ \cline{2-4} 
 & \textbf{ARR@1} & \textbf{ARR@5} & \textbf{ARR@10} \\ \hline
Alexnet layer 6+ SSF + KSVD & \textbf{1} & \textbf{3.4} & \textbf{5.3} \\
LGHP & 0.4 & 1 & 1.6 \\
LDP & 0.25 & 0.75 & 1.2 \\
LTrP & 0.2 & 0.5 & 1 \\
LVP & 0.3 & 0.8 & 1.4 \\ \hline
\textbf{} & \multicolumn{3}{c}{\textbf{Gallagher(\%)}} \\ \cline{2-4} 
\textbf{} & \textbf{ARR@1} & \textbf{ARR@3} & \textbf{ARR@5} \\ \hline
Alexnet layer 6+ SSF + Kmeans & 1.1 & 3.2 & 5 \\
LGHP & \textbf{4.1} & \textbf{8} & \textbf{9.5} \\
LDP & 2.9 & 5.1 & 7 \\
LTrP & 1.6 & 3 & 3.5 \\
LVP & 3.1 & 6.5 & 8.3 \\ \hline
\end{tabular}
\end{table}

Figures 2 - 3 illustrate the Precision-Recall (PR) and cumulative matching characteristic (CMC) curves of the face retrieval experiments. These figures show PR and CMC curves on the evaluated datasets using all the tested CNN models with their different layers with $K$-Means and $K$-SVD and SSF.
\begin{figure}[ht]
\centering     
\subfigure[]{
\label{fig:a}
\includegraphics[width=60mm]{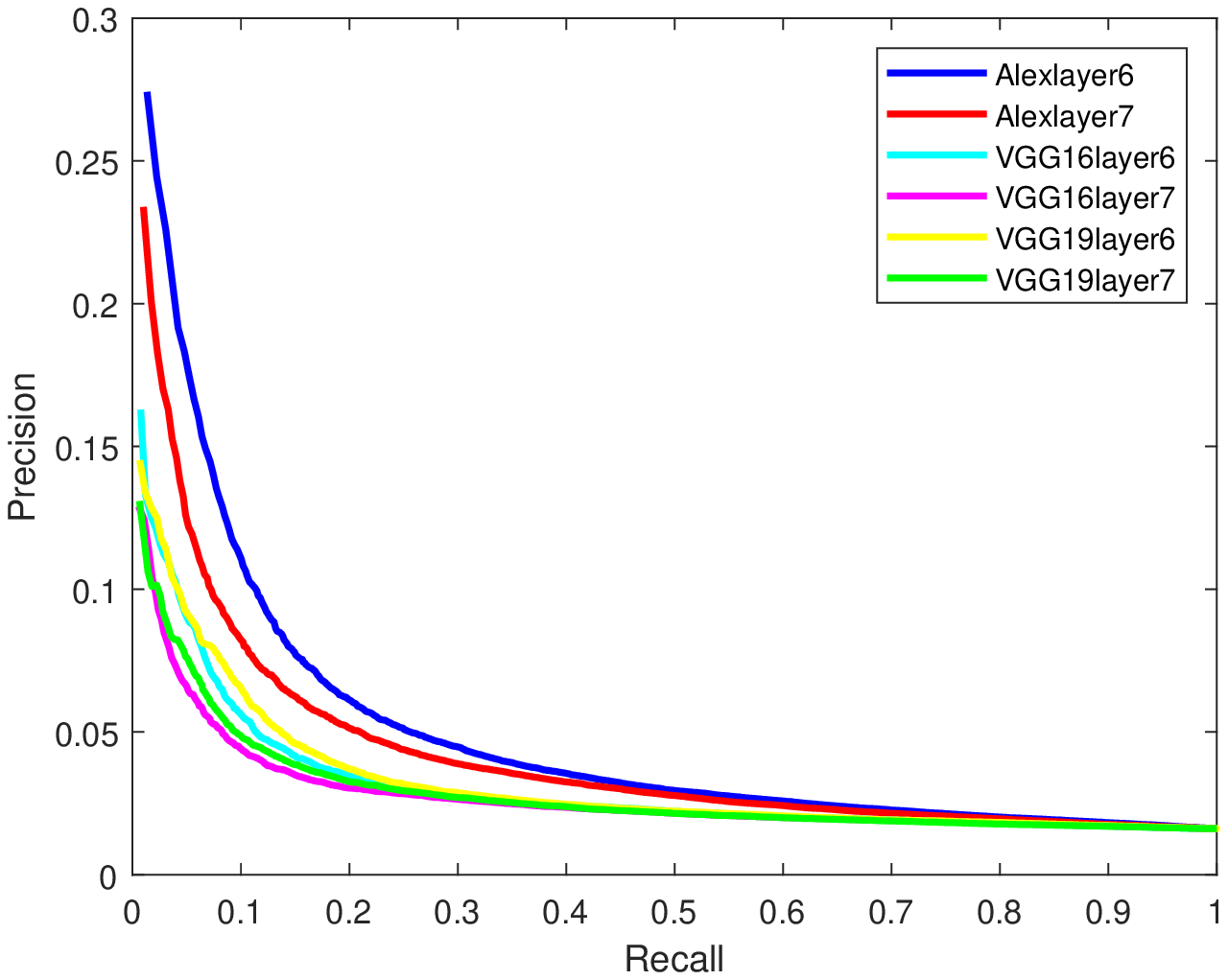}}
\subfigure[]
{\label{fig:b}
\includegraphics[width=60mm]{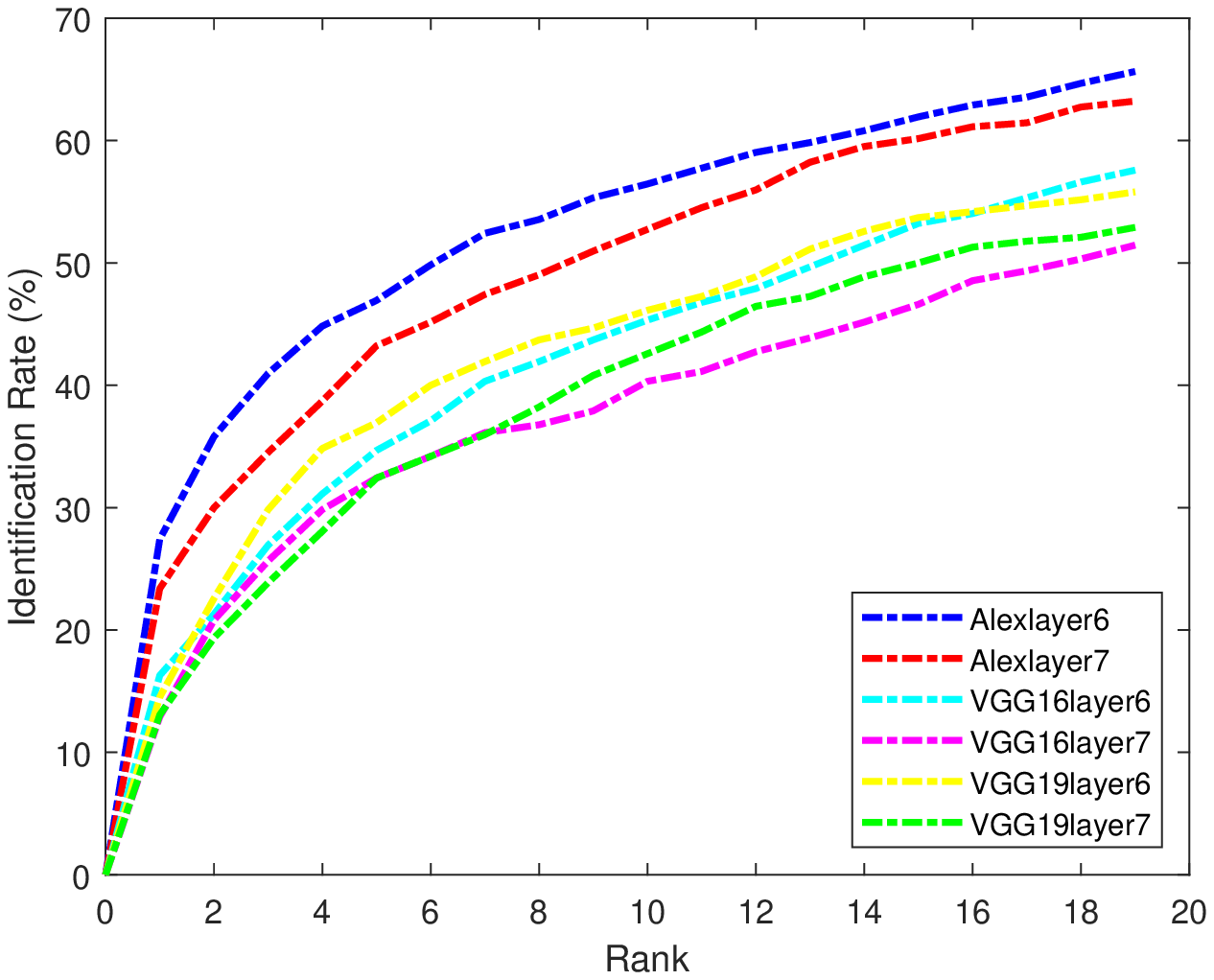}}

\subfigure[]
{\label{fig:b}
\includegraphics[width=60mm]{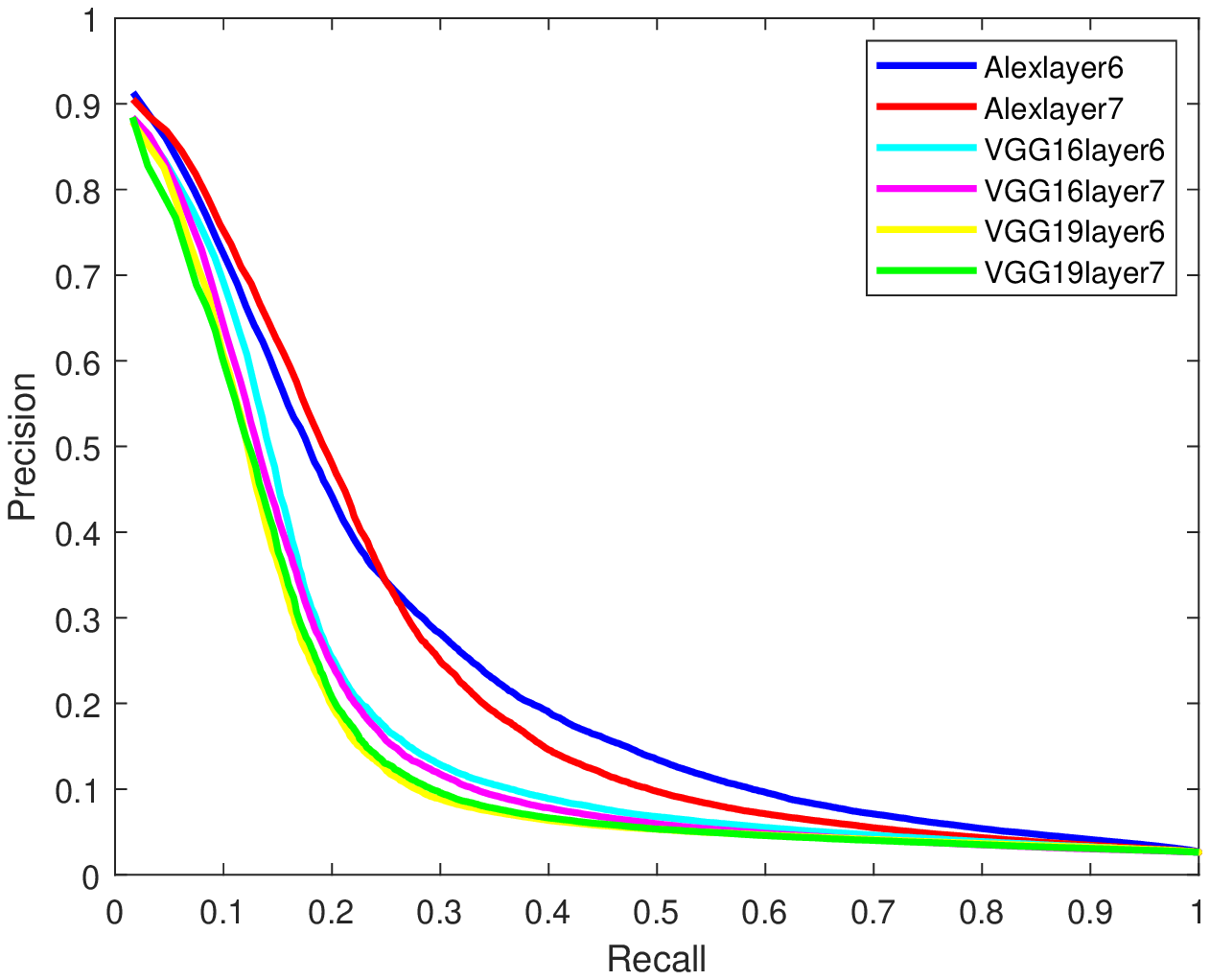}}
\subfigure[]
{\label{fig:b}
\includegraphics[width=60mm]{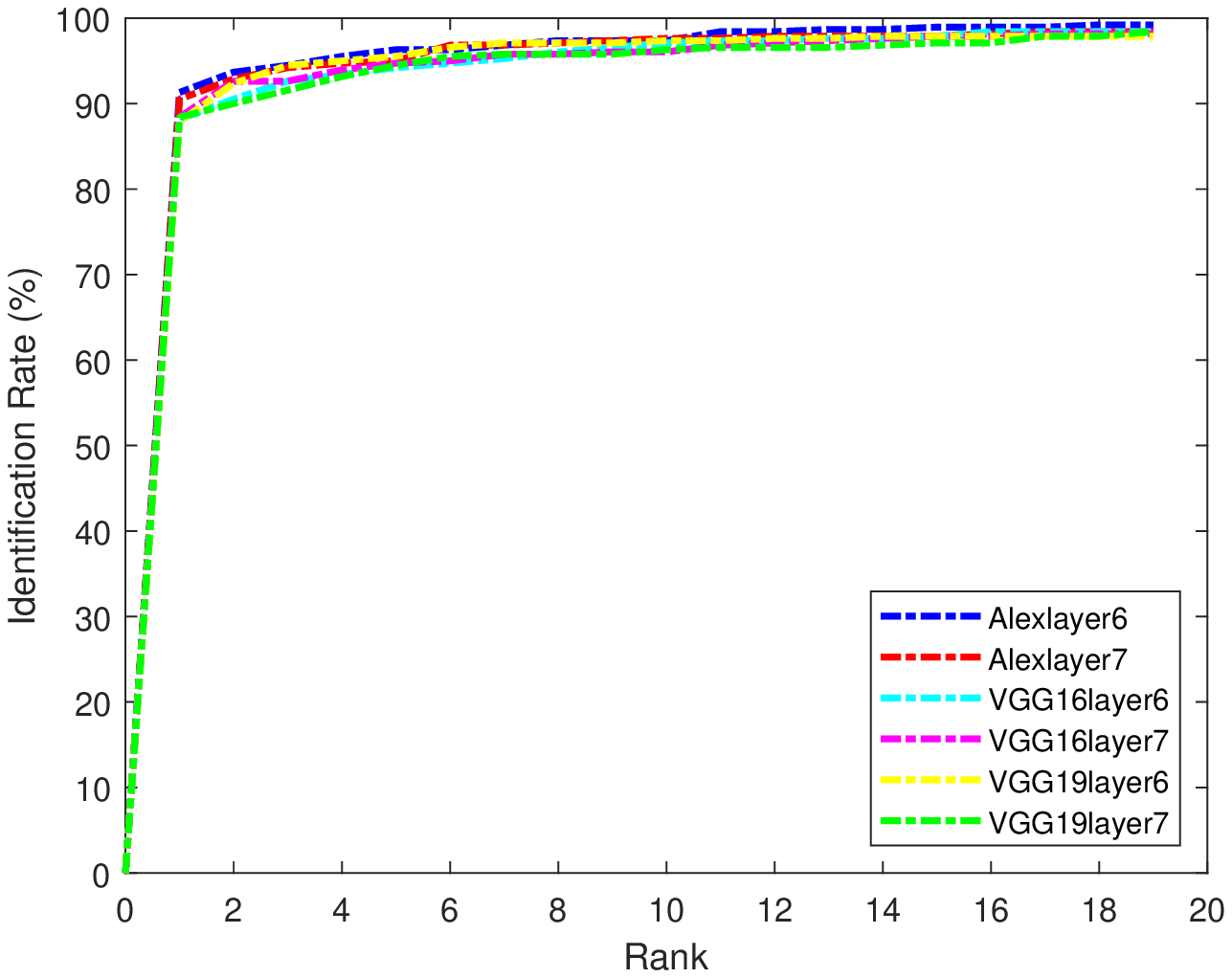}}

\subfigure[]
{\label{fig:b}
\includegraphics[width=60mm]{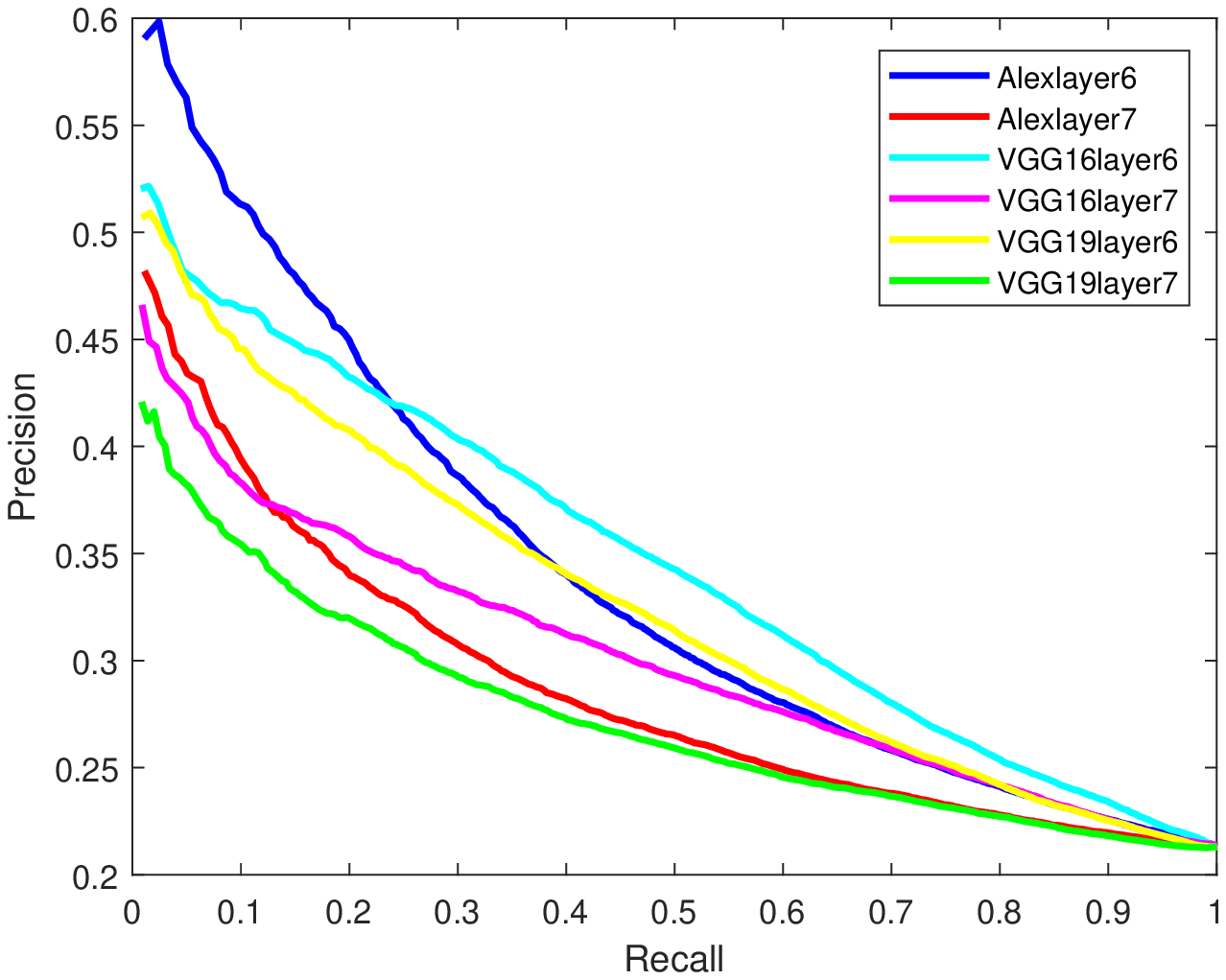}}
\subfigure[]
{\label{fig:b}
\includegraphics[width=60mm]{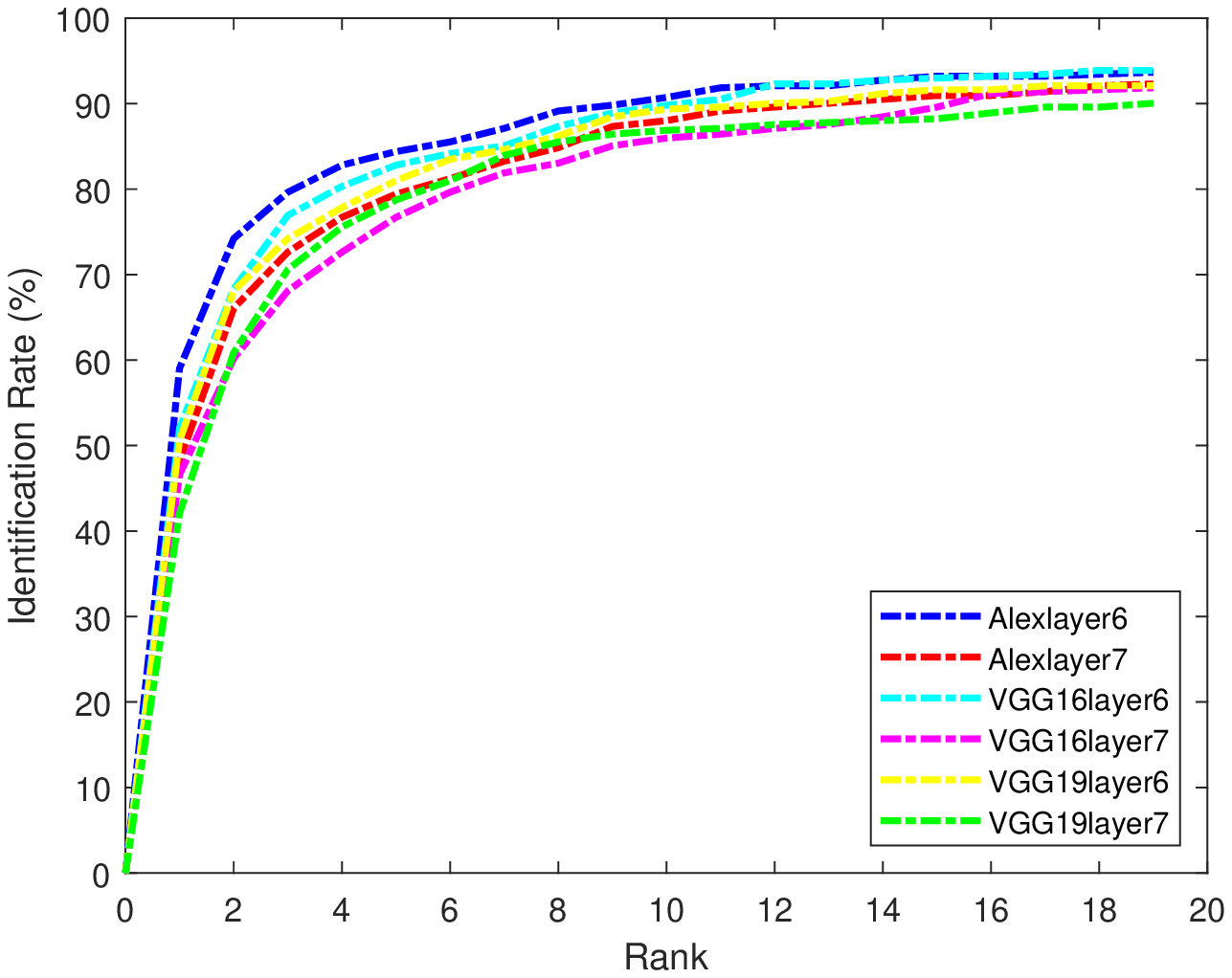}}
\caption{PR (Left) and CMC (right) curves of the $K$-Means with SSF results on LFW (a-b), Yal-b (c-d) and Gallagher (e-f)}
\end{figure}


\begin{figure}[ht]
\centering     
\subfigure[]{
\label{fig:a}
\includegraphics[width=60mm]{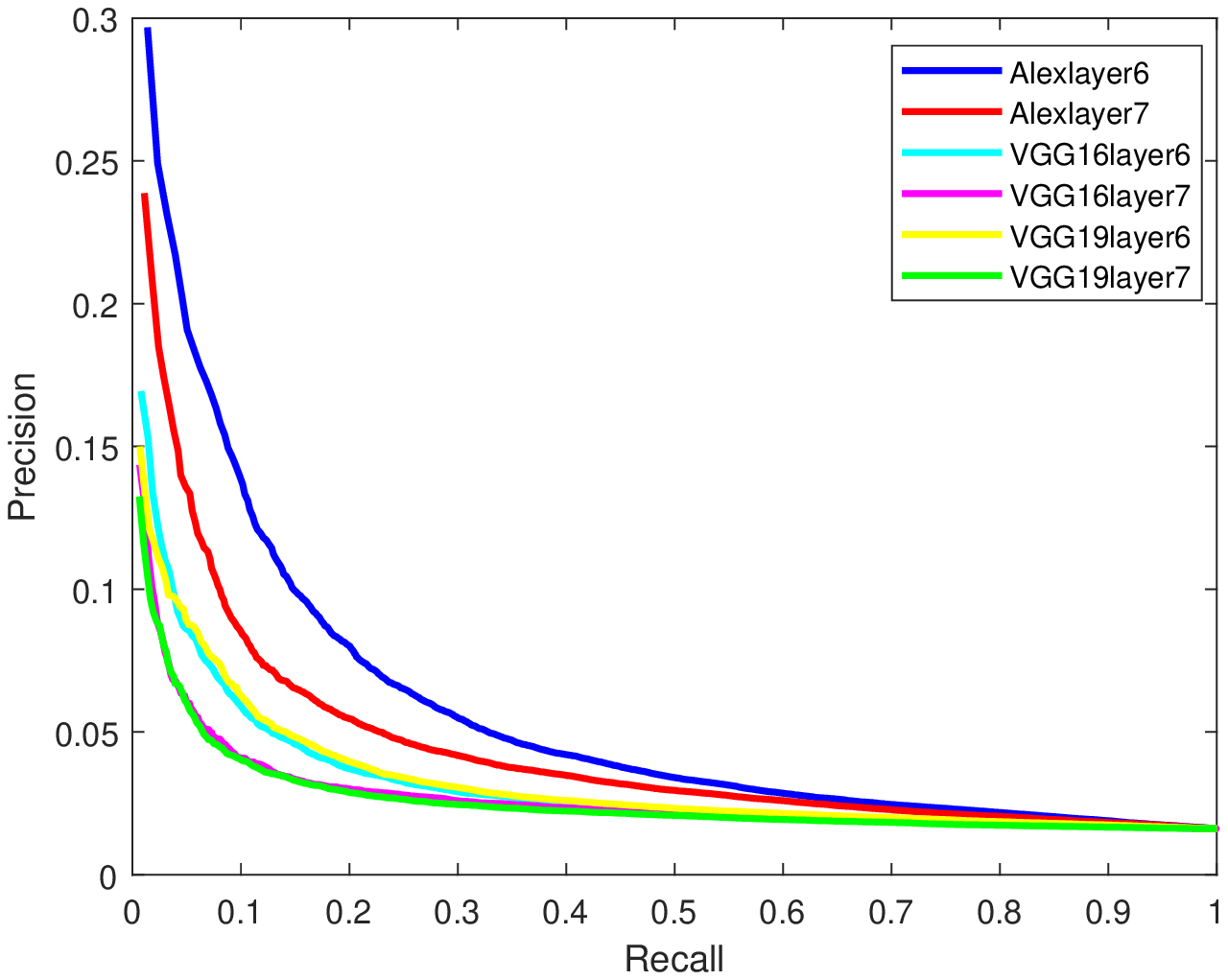}}
\subfigure[]
{\label{fig:b}
\includegraphics[width=60mm]{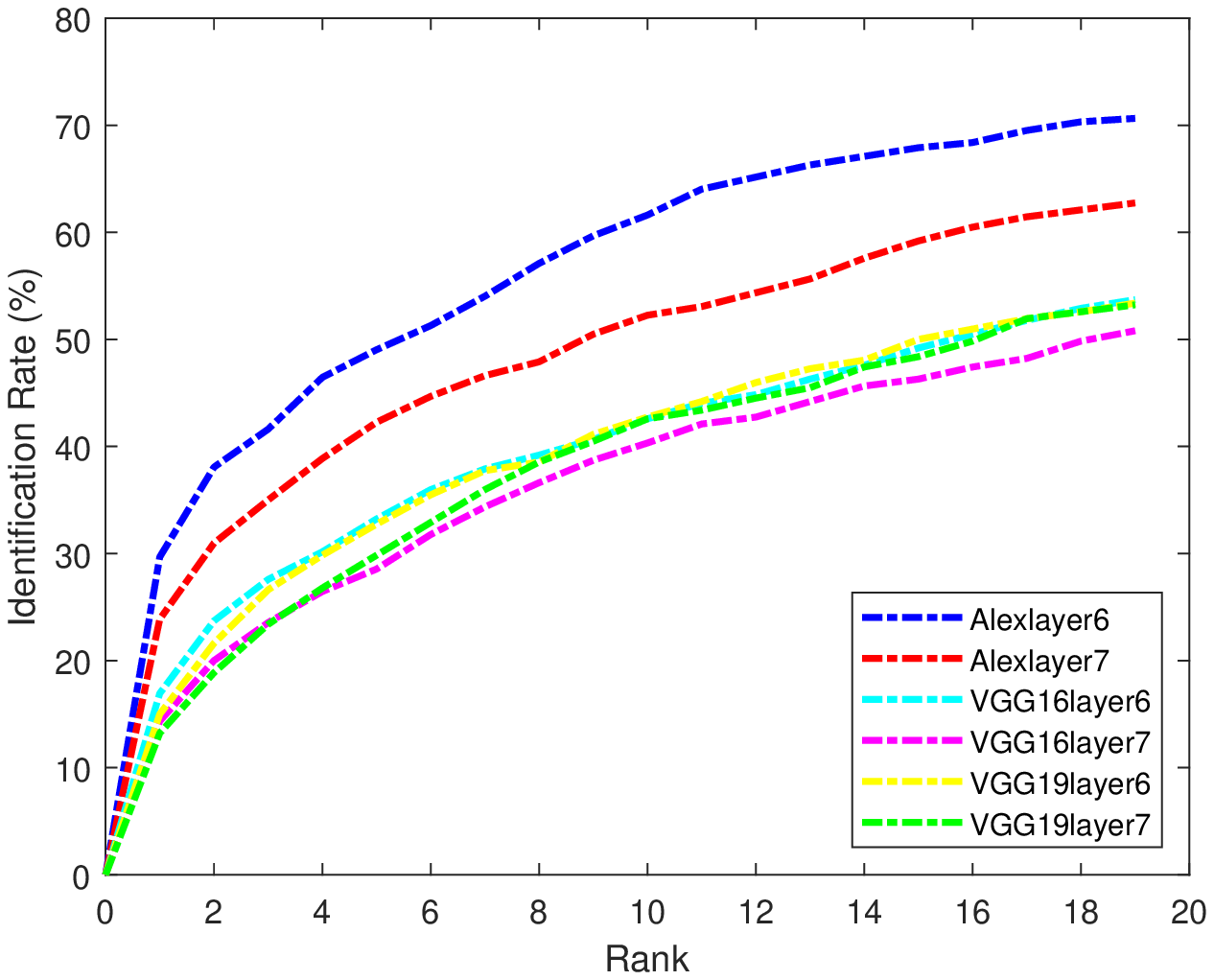}}

\subfigure[]
{\label{fig:b}
\includegraphics[width=60mm]{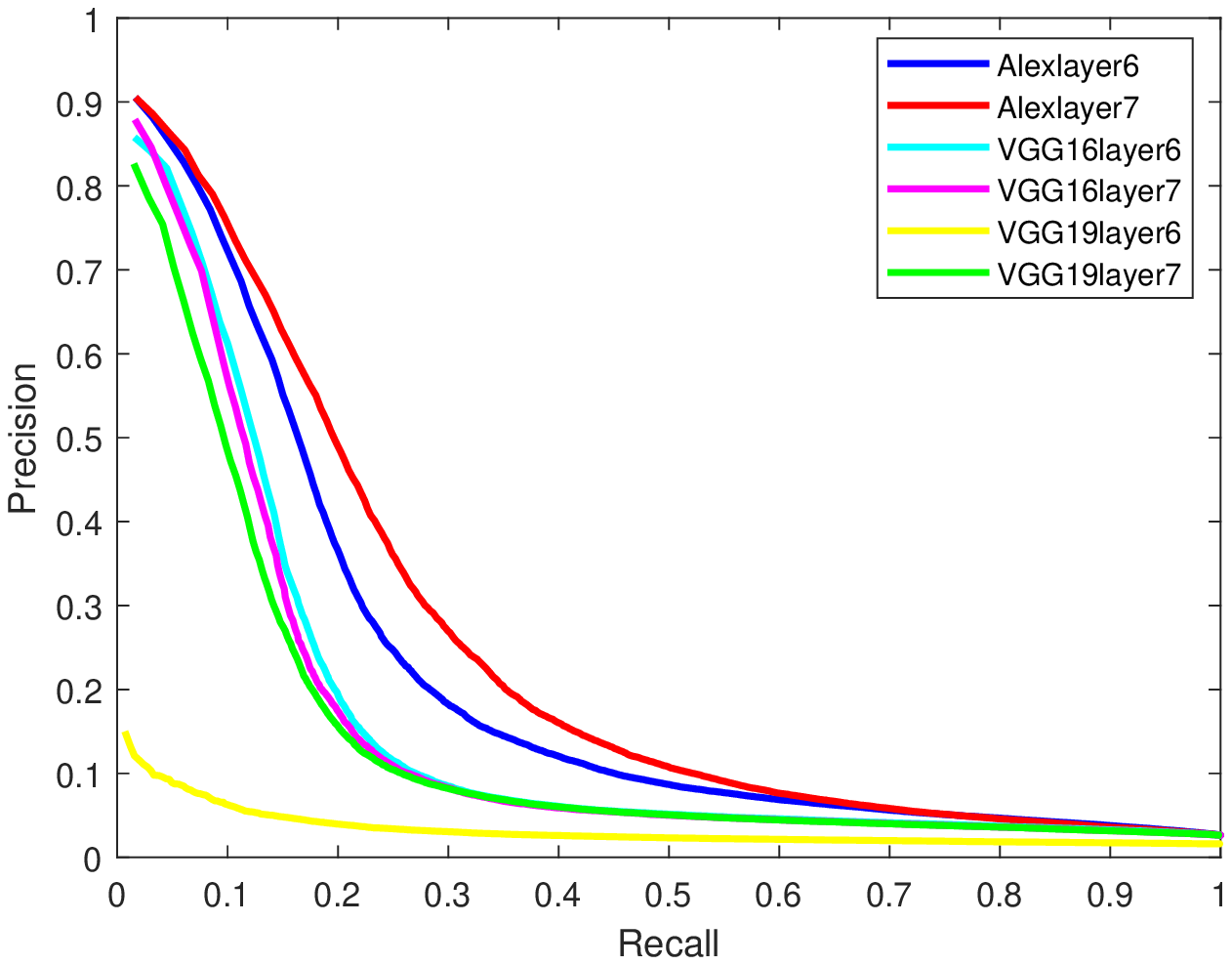}}
\subfigure[]
{\label{fig:b}
\includegraphics[width=60mm]{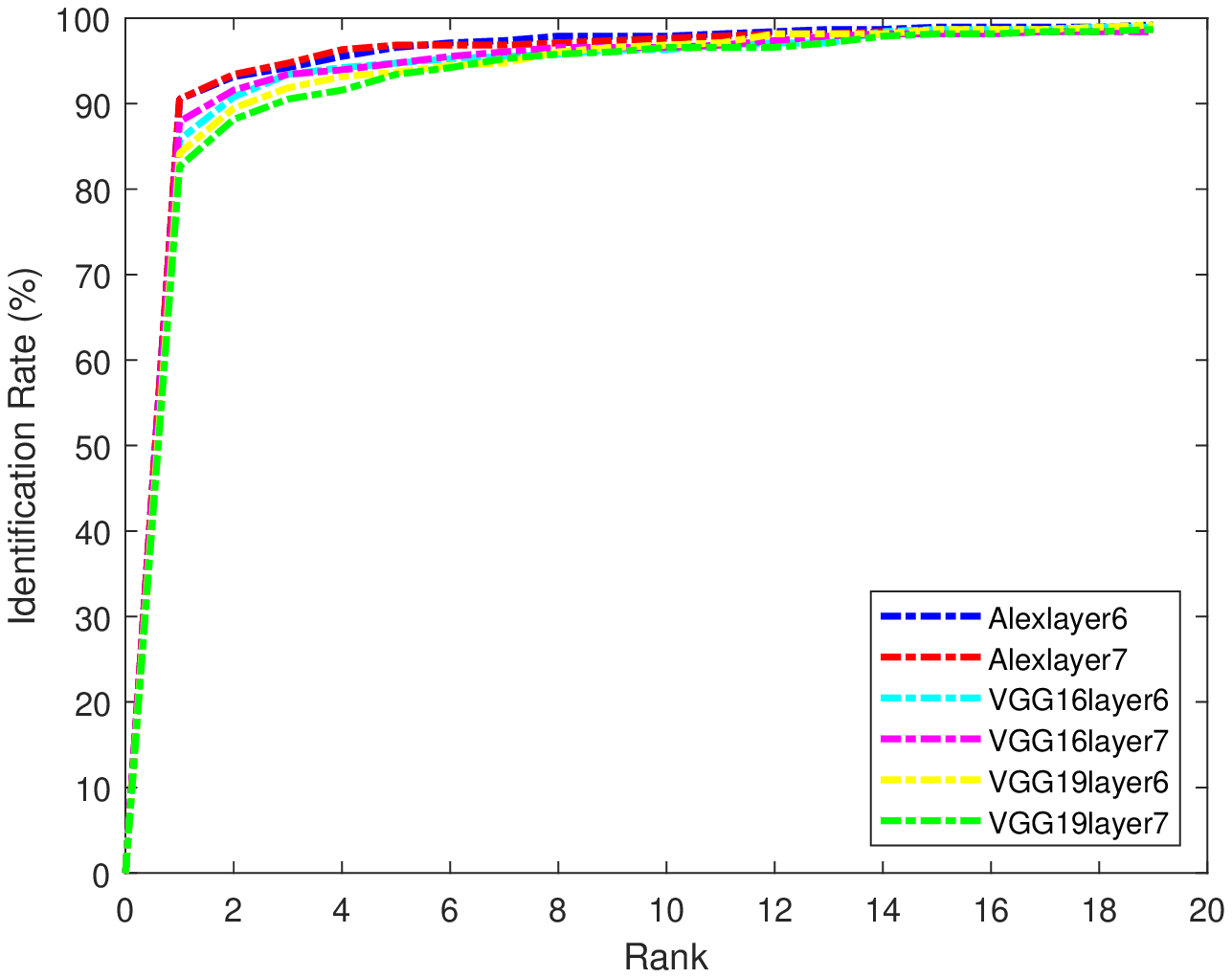}}

\subfigure[]
{\label{fig:b}
\includegraphics[width=60mm]{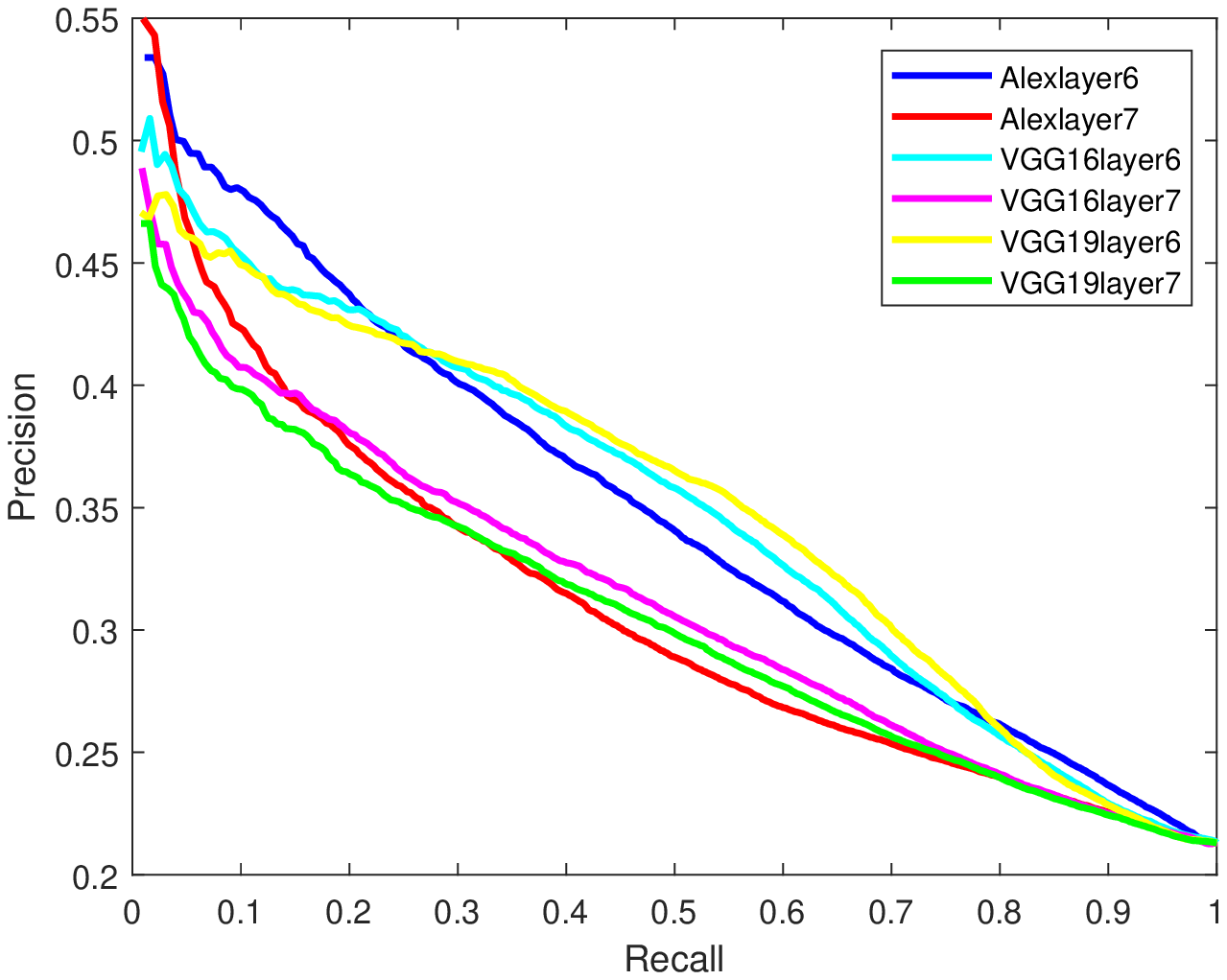}}
\subfigure[]
{\label{fig:b}
\includegraphics[width=60mm]{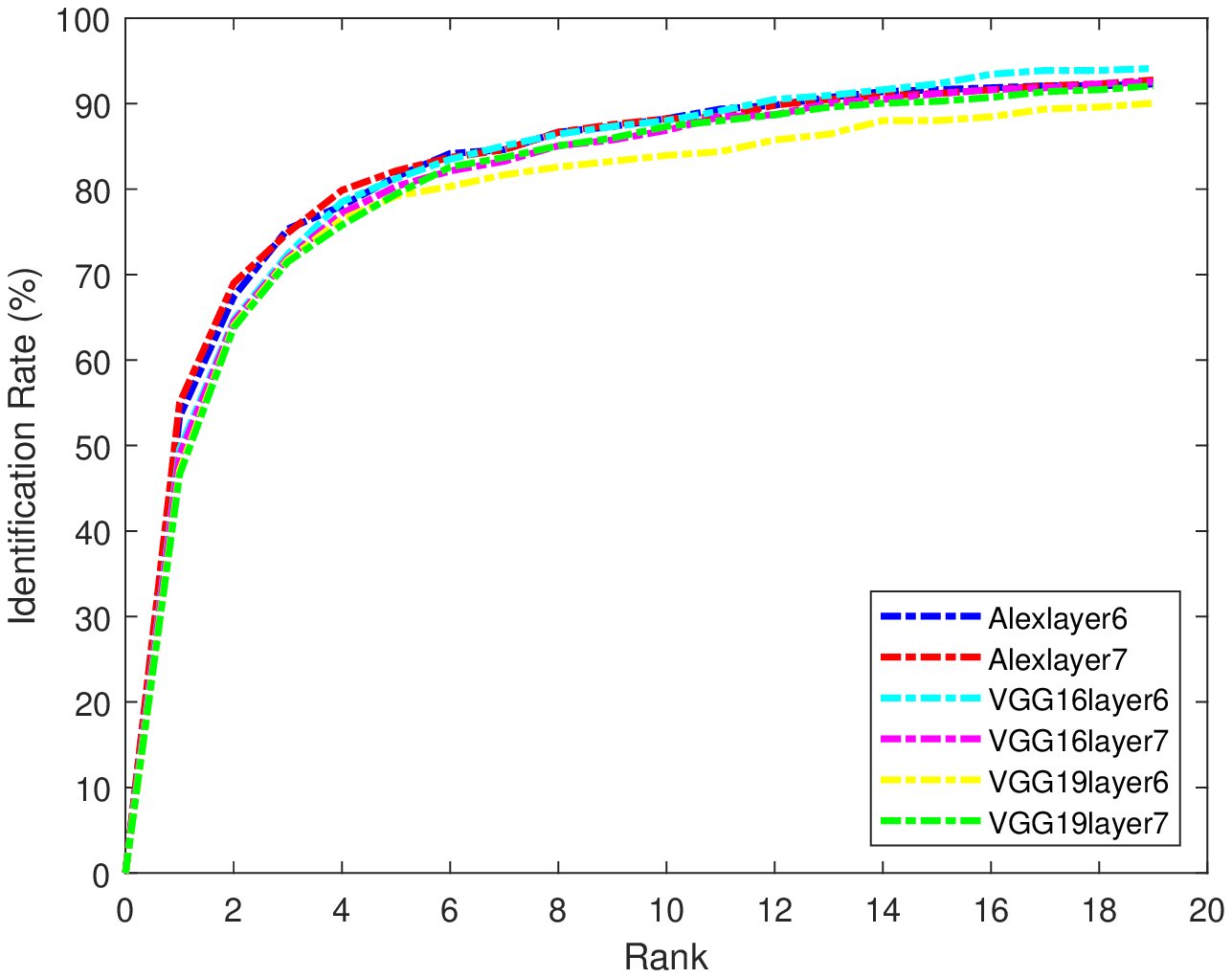}}
\caption{PR (Left) and CMC (right) curves of the $K$-SVD with SSF results on LFW (a-b), Yal-b (c-d) and Gallagher (e-f)}
\end{figure}

\clearpage
As it can be seen from the above figures, in terms of dictionaries, $K$-SVD performs better than $K$-means in most cases. AlexNet obtains better precision than the very deep models. This is due to its better utilization ability for the parameters, especially, for the full face features. In addition, the top matches (CMC) (at the right of Figures 3-7) are identified correctly and more accurately on the Yale-B dataset, where the retrieval rate starts to approach 100\% at rank 13. On the other hand, the retrieval rates are lower on LFW. Alexlayer6 with $K$-SVD and SSF obtains the best retrieval rate at rank 14, which is very close to 70\%. This is followed by the Alexlayer7 as it achieves  60\% at the same rank. The other models with different layers achieve almost 50\% at the same rank. The low retrieval rates of the LFW is probably due to the small number of images left for the training data.

Figure 4 presents the PR curves for the best retrieval results on all datasets. The left side of each figure illustrates the PR curves on the whole dataset, while the right one shows the precision as a function of the top matches.

\begin{figure}[ht]
\centering     
\subfigure[]{
\label{fig:a}
\includegraphics[width=60mm]{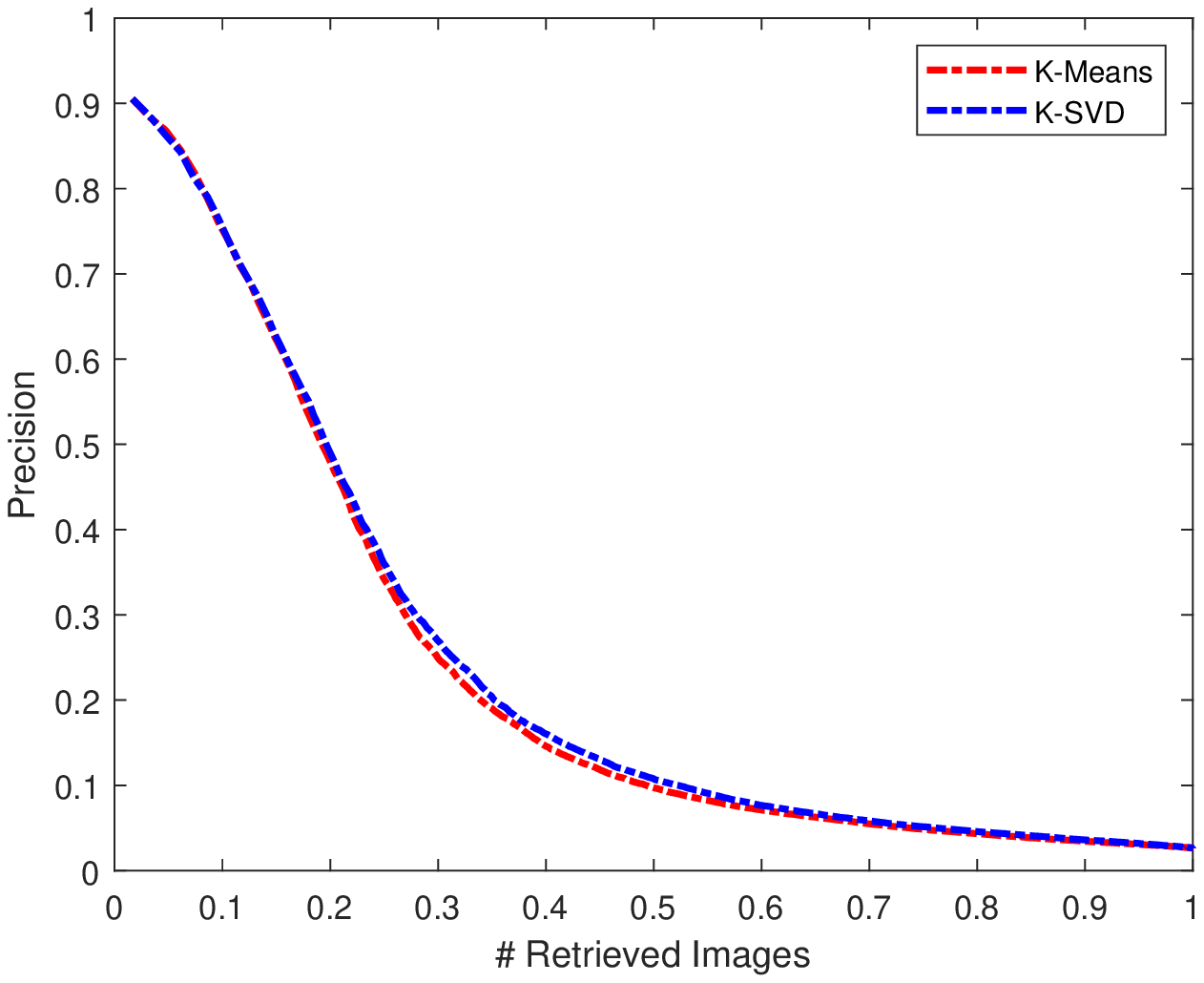}}
\subfigure[]
{\label{fig:b}
\includegraphics[width=60mm]{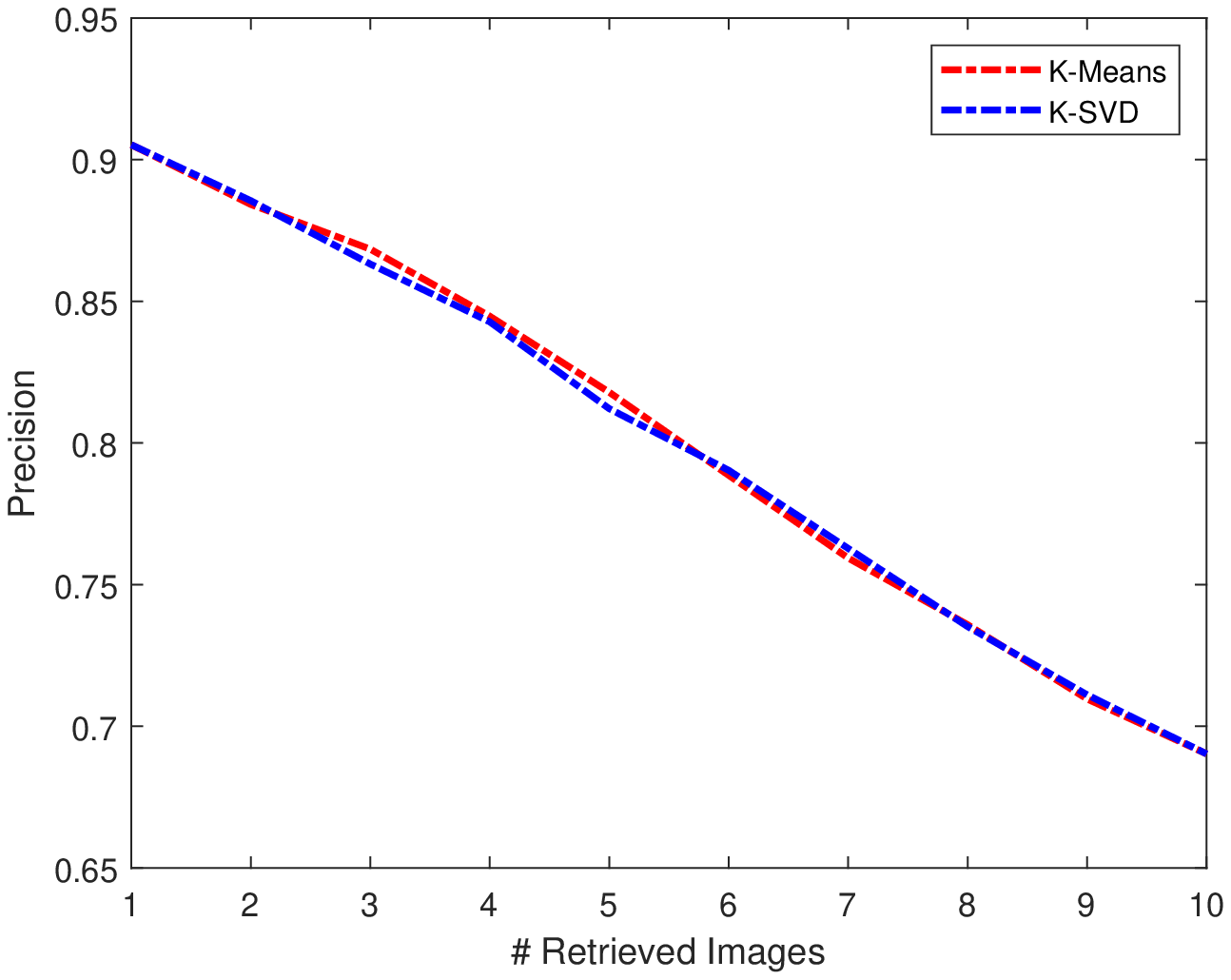}}

\subfigure[]
{\label{fig:b}
\includegraphics[width=60mm]{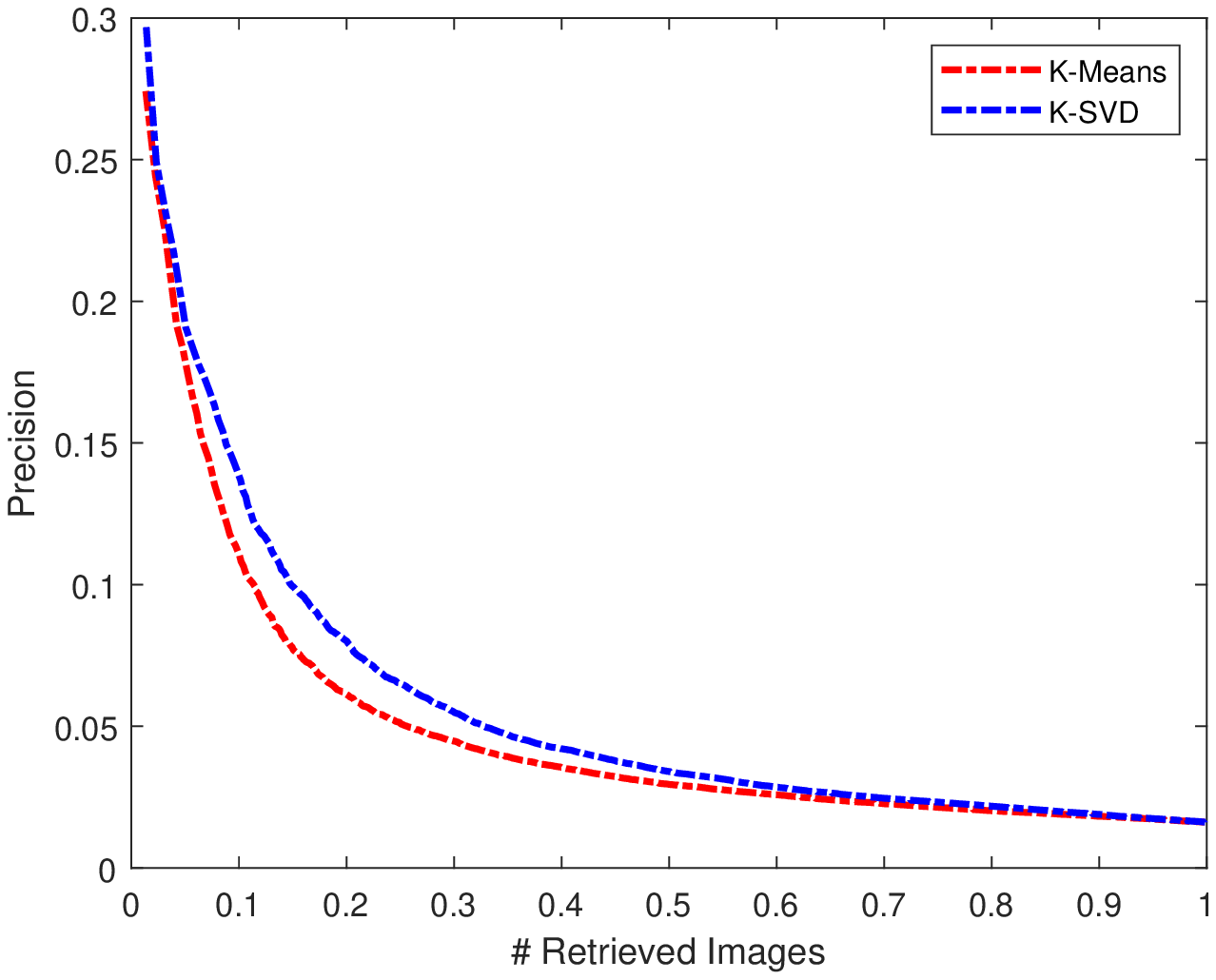}}
\subfigure[]
{\label{fig:b}
\includegraphics[width=60mm]{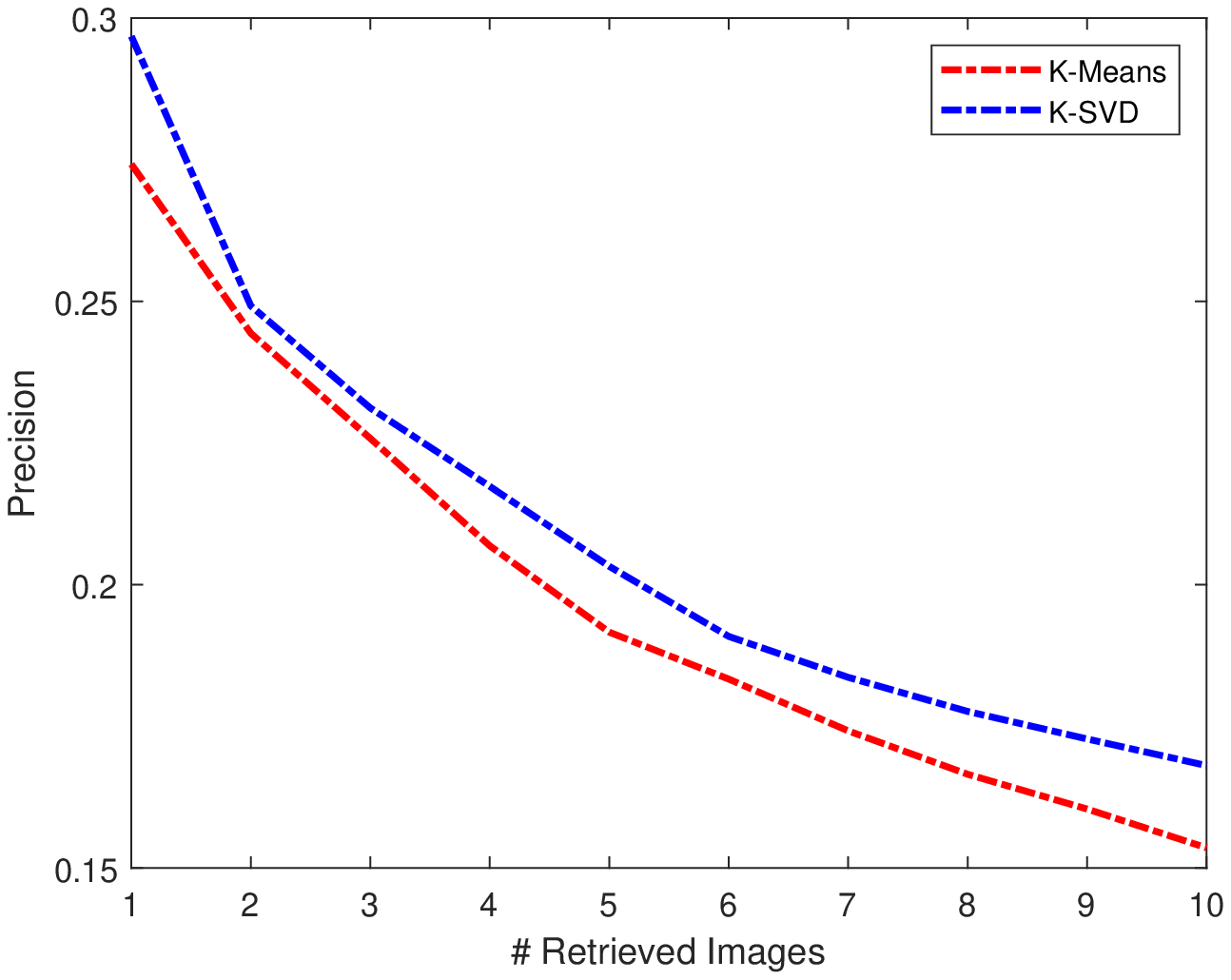}}

\subfigure[]
{\label{fig:b}
\includegraphics[width=60mm]{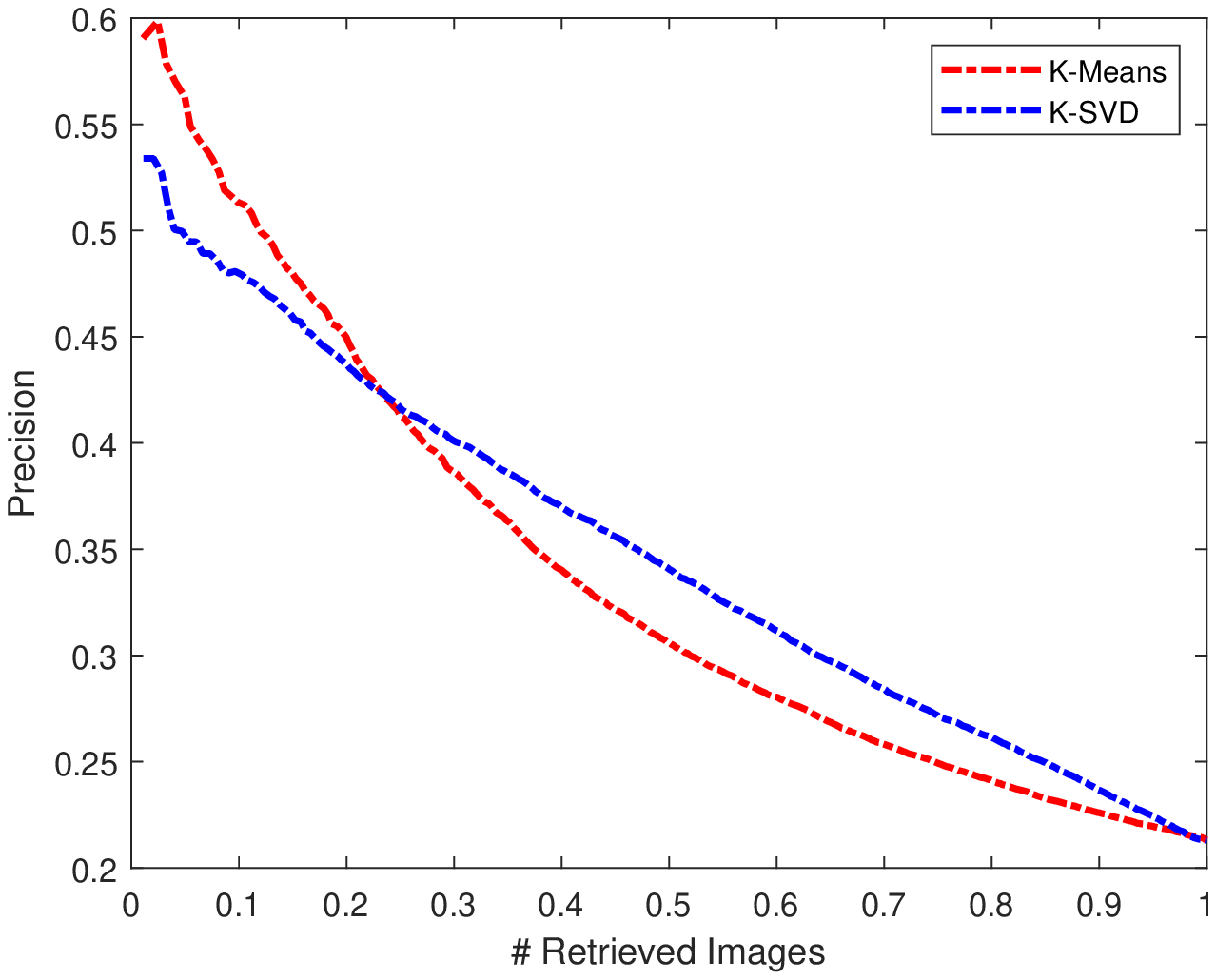}}
\subfigure[]
{\label{fig:b}
\includegraphics[width=60mm]{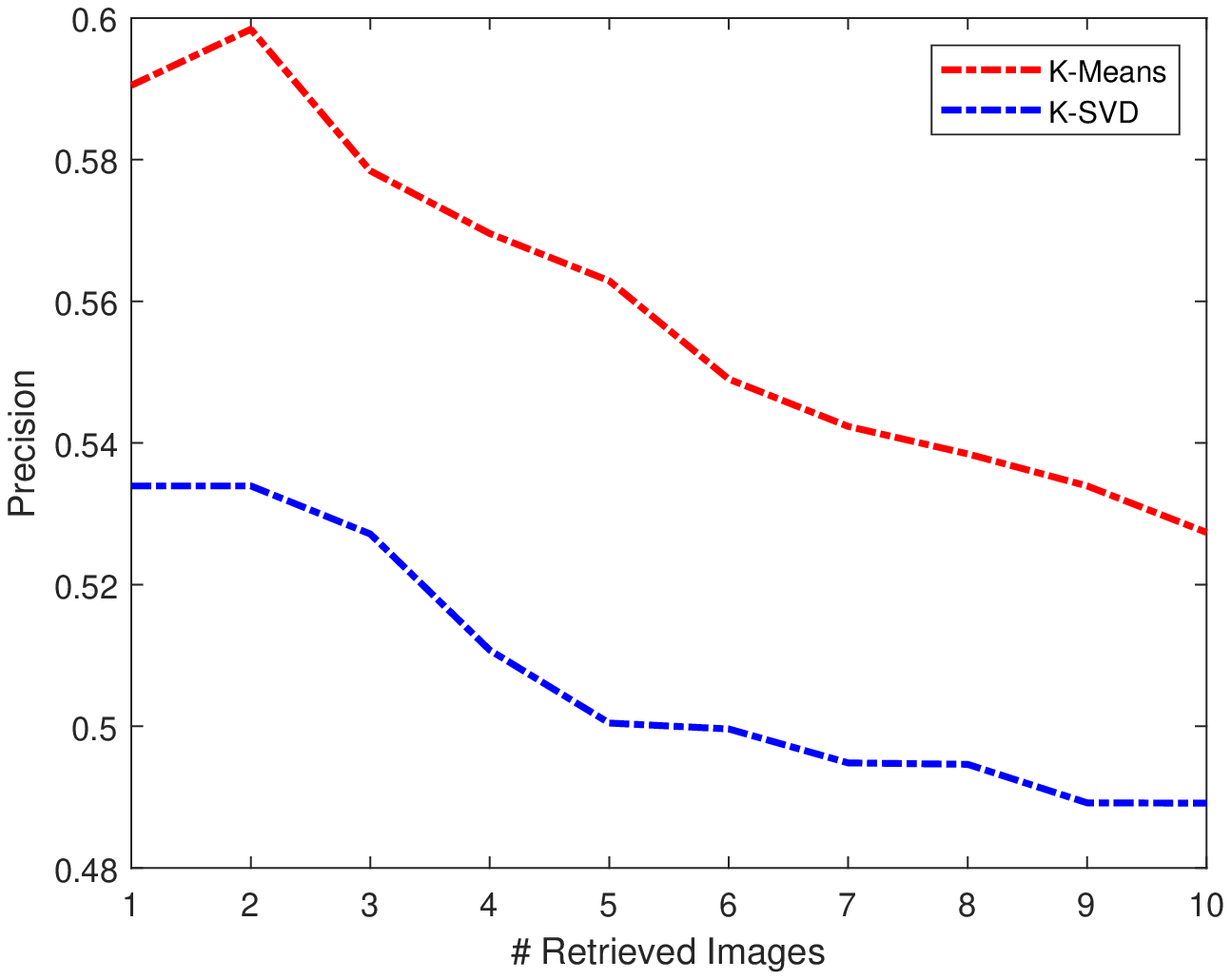}}
\caption{PR (Left) and precision with top matches (right) curves of the performance of $K$-SVD compared to $K$-Means with SSF using the best features on the three datasets results on Yal-b (a-b), LFW (c-d) and Gallagher (e-f)}
\end{figure}

As one can see in Figure 4, the $K$-SVD performs better than the $K$-means with SSF on LFW dataset, whereas both dictionaries perform almost the same on the Yale-B dataset. This is also the case for Gallagher dataset as $K$-SVD results are superior to $K$-means on average. For the Yale-B dataset, the retrieval precision starts from 91\% and continues to decrease reaching 70\% when 10 images are retrieved. This dramatic decrease in the retrieval precision can be attributed to the nature of the images in the Yale-B dataset having dark images and various facial expressions. The situation is similar for LFW; however, the latter has more challenging images with various backgrounds in addition to different hair style, cloths and facial expressions. 

Localizing the face within these images may improve the retrieval results. However, localizing the faces in order to focus on the main features of the face in the LFW dataset is out of the scope of this paper and may be taken up as a future work. In the Gallagher dataset, the faces are localized but the pose varies significantly; in addition, this dataset poses the same challenges as the previous datasets, including varying illumination and real-life facial expressions.

\clearpage

\section{Conclusion}

This paper investigates the use of different deep learning models for face image retrieval, namely, AlexNet FC6, AlexNet FC7, VGG16layer6, VGG16layer7, VGG19layer6, and VGG19layer7. The models utilize two types of dictionary learning techniques, $K$-means and $K$-SVD, in addition to the use of some coefficient learning techniques such as the Homotopy, Lasso, Elastic Net and SSF. The comparative results of the experiments conducted on three standard challenging face image datasets show that the best performers for face image retrieval are Alexlayer7 with $K$-means and SSF, Alexlayer6 with $K$-SVD and SSF, and Alexlayer6 with $K$-means and SSF. The APR and ARR of these methods were further compared to some of the local descriptor-based methods found in the literature. The experimental results show that the deep learning approaches outperform most of the methods compared, and therefore they can be recommended for practical use in face image retrieval.

Despite the good performance of the deep features, the retrieval process is still not perfect. In particular, when tested on non-cropped face images, such imperfect performance might be attributed to several challenges posed by the images found in the datasets we used, namely, (1) complex and different backgrounds; (2) darknes of the images; and (3) different facial expressions. The first problem can be solved by localizing the faces in order to focus on the main features of the face rather than the complex background; this can be done efficiently using the method proposed in \cite{hassanat2016color}. The second problem can be solved using image enhancement in a preprocessing stage. Finally, the third problem can be solved using some transformation of the face image to alleviate the differences in facial expressions of the same subject; this can be done using the method proposed in \cite{hassanat2018magnetic}. We plan to include these components in our methodology in future. Our future works will also include the use of deep features with dictionary learning to solve other relevant problems handled in  \cite{hassanat2018identifying}, \cite{hassanat2017victory} and \cite{hassanat2017classification}. We also plan to increase the speed of the retrieval process using efficient indexing techniques, such as, \cite{hassanat2018furthest}, \cite{hassanat2018furthestDT} and \cite{hassanat2018norm}.

\section*{Acknowledgements}
The first author would like to thank Tempus Public Foundation for sponsoring his PhD study, also, this paper is under the project EFOP-3.6.3-VEKOP-16-2017-00001 (Talent Management in Autonomous Vehicle Control Technologies), and supported by the Hungarian Government and co-financed by the European Social Fund.

\bibliographystyle{unsrtnat}
\bibliography{Ref}

\end{document}